\documentclass{article}

    \PassOptionsToPackage{numbers, compress}{natbib}

    \usepackage[preprint]{neurips_2024}

\usepackage[utf8]{inputenc} 
\usepackage[T1]{fontenc}    
\usepackage{hyperref}       
\usepackage{url}            
\usepackage{booktabs}       
\usepackage{amsfonts}       
\usepackage{nicefrac}       
\usepackage{microtype}      
\usepackage{xcolor}         
\usepackage{stix}

\usepackage{mathtools,xparse}

\usepackage{ulem}
\usepackage{multirow}
\usepackage{makecell}
\usepackage{tabulary}
\usepackage{booktabs}
\usepackage{bm}
\usepackage {wrapfig} 

\usepackage{pifont}
\usepackage{subfigure}
\usepackage{wrapfig}
\usepackage{bbding}
\usepackage{colortbl} 
\usepackage{ulem}

\newcommand{\tabincell}[2]{\begin{tabular}{@{}#1@{}}#2\end{tabular}}
\newcommand{\dt}{\Delta}

\title{Heracles: A Hybrid SSM-Transformer Model for High-Resolution Image  and Time-Series Analysis}

\author{Badri N. Patro\\
Microsoft\\
{\tt\small badripatro@microsoft.com}\\
\and 
Suhas Ranganath\\
Microsoft\\
{\tt\small sranganath@microsoft.com}\\
\and 
Vinay P. Namboodiri\\
University of Bath\\
{\tt\small vpn22@bath.ac.uk} \\
\and
Vijay S. Agneeswaran\\
Microsoft\\
{\tt\small vagneeswaran@microsoft.com}
}

\begin{document}

\maketitle
\vspace{-0.3cm}

\begin{abstract}
Transformers have revolutionized image modeling tasks with adaptations like DeIT, Swin, SVT, Biformer, STVit, and FDVIT. However, these models often face challenges with inductive bias and high quadratic complexity, making them less efficient for high-resolution images. State space models (SSMs) such as Mamba, V-Mamba, ViM, and SiMBA offer an alternative  to handle high resolution images in computer vision tasks. These SSMs encounter two major issues. First, they become unstable when scaled to large network sizes. Second, although they efficiently capture global information in images, they inherently struggle with handling local information. To address these challenges, we introduce Heracles, a novel SSM that integrates a local SSM, a global SSM, and an attention-based token interaction module. Heracles leverages a Hartely kernel-based state space model for global image information, a localized convolutional network for local details, and attention mechanisms in deeper layers for token interactions. Our extensive experiments demonstrate that Heracles-C-small achieves state-of-the-art performance on the ImageNet dataset with 84.5\% top-1 accuracy. Heracles-C-Large and Heracles-C-Huge further improve accuracy to 85.9\% and 86.4\%, respectively. Additionally, Heracles excels in transfer learning tasks on datasets such as CIFAR-10, CIFAR-100, Oxford Flowers, and Stanford Cars, and in instance segmentation on the MSCOCO dataset. Heracles also proves its versatility by achieving state-of-the-art results on seven time-series datasets, showcasing its ability to generalize across domains with spectral data, capturing both local and global information. The project page is available at this link.\url{https://github.com/badripatro/heracles}

\end{abstract}

\section{Introduction}
\label{sec:intro}

\begin{figure}[thb]%
\centering
\includegraphics[width=0.849\textwidth]{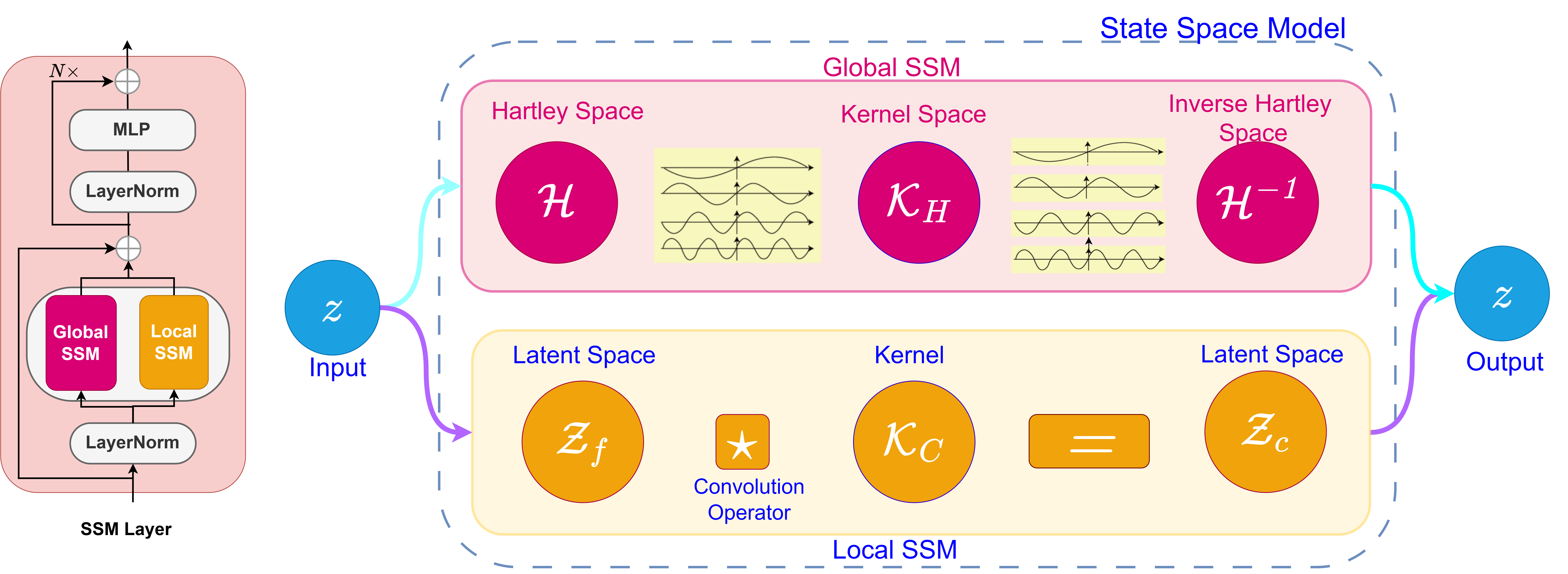}
\caption{Heracles uses Global SSM using Hartley transform and Learnable Kernel for Capturing Global Features, Coupled with Local SSM Utilizing ConvNet for Precise Localization.}\label{fig:intro}
\vspace{-0.3in}
\end{figure}

Transformers like the Vision Transformer (ViT) have shown promising results in computer vision tasks such as image recognition, object detection, and semantic segmentation. The transformer architecture serves as the backbone of modern large vision models (LVMs) and even smaller vision models. However, it has been demonstrated in the literature that transformers struggle with high-resolution image data, which involve a large number of patch tokens or long sequences, primarily due to their \(O(N^2)\) complexity and lack of inductive bias \cite{liu2024vmamba}\footnote{In VMamba paper, for example, includes a performance comparison diagram (Figure 11) showing that transformer performance declines as image resolution increases}. Several adaptations have been proposed to improve transformer performance on these tasks, including DeIT \cite{touvron2021training}, Flatten Transformer \cite{han2023flatten}, MixFormer \cite{CuiMixFormer24}, SVT \cite{patro2023scattering}, Biformer \cite{zhu2023biformer}, STVit \cite{huang2023vision}, and FDVIT \cite{xu2023fdvit}. Despite these advancements, these models still face issues with quadratic complexity and lack of inductive bias, especially for high-resolution images. For example, with an image resolution of 224x224 pixels and a typical patch size of 16x16 pixels, there are 196 patches or sequences. As image resolution increases, the number of sequences increases significantly, and transformer complexity grows by \(O(N^2)\).

State Space Models (SSMs) have emerged as a promising alternative to transformers for long sequence modeling, beginning with the introduction of S4 \cite{gu2021efficiently}. The Structured State Space sequence model (S4) is designed to capture long dependencies within sequences using a continuous-time state space model. S4 was the first state space model to solve the path-X vision task in the LRA benchmark and reduces computational complexity to \(O(N\log(N))\). Several variants of S4 have been developed, including High-Order Polynomial Projection Operators (HiPPO) \cite{albert2020hippo}, S4nd \cite{nguyen2022s4nd}, Hyena \cite{poli2023hyena}, Diagonal State Spaces (DSS) \cite{gupta2022diagonal}, Gated State Spaces (GSS) \cite{mehta2022long}, Linear Recurrent Unit (LRU) \cite{orvieto2023resurrecting}, Liquid-S4 \cite{hasani2021liquid}, and Mamba \cite{gu2023mamba}. In the vision domain, various Mamba variants have also been proposed, including Vision Mamba \cite{zhu2024vision}, V-Mamba\cite{liu2024vmamba}, and SiMBA \cite{patro2024simba}.

V-Mamba \cite{liu2024vmamba}, Vision Mamba \cite{zhu2024vision}, and SiMBA \cite{patro2024simba} utilize vanilla Mamba to tackle computer vision tasks. However, V-Mamba and Vision Mamba exhibit a performance gap compared to state-of-the-art transformer models such as SpectFormer \cite{patro2023spectformer}, SVT \cite{patro2023scattering}, BiFormer \cite{zhu2023biformer}, STViT \cite{huang2023vision}, FDViT \cite{xu2023fdvit}, MixMAE \cite{liu2023mixmae}, and Flatten Transformer \cite{han2023flatten}, as shown in \cite{liu2024vmamba}. We hypothesize that this gap is due to their inability to capture local information in images, as SSMs typically capture only global information. Some transformers, like SVT and LGViT \cite{Zhou2023LGViT}, are designed to capture local information but are computationally expensive, making them unsuitable for long sequences. While Mamba inherently captures global information and some local information through convolutional layers, it suffers from stability issues in larger networks, leading to gradient explosion or vanishing, resulting in NAN values during training for vision tasks. SiMBA partially addresses this issue using EinFFT \cite{patro2024simba} but does not explicitly capture local information.

This paper presents a novel approach to address the instability issues in vanilla Mamba and the inability of SSMs to capture local information by designing appropriate kernels using real-valued transforms. We propose Heracles, a new SSM that employs real-valued transforms such as Discrete Cosine Transforms (DCT) and Hartley Transforms to capture global information, along with convolutional networks to explicitly capture local information. Heracles uses a simplified kernel that is Hartley Kernel inspired by Diagonal State Spaces (DSS) \cite{gupta2022diagonal}. Heracles uses SSM (Hartley) and convolutional networks in a parallel fashion. It is distinct from the vanilla Mamba architecture and results in efficiencies. Additionally, Heracles incorporates multi-head attention networks in its deeper layers, which helps capture long-range dependencies between patches. The architecture of Heracles is depicted in Figure \ref{fig:intro}. We believe that this also enables Heracles to achieve state-of-the-art performance on both image and time series datasets. Given the similar spectral properties and the need to capture both local and global information, this discussion is also applicable to time series data. We demonstrate the effectiveness of Heracles by evaluating its performance on six time series datasets.




Our contribution is as follows:
\begin{itemize}
    \item \textbf{Heracles}: We introduce Heracles, an innovative SSM that integrates a Hartely kernel-based state space model for global image information with a localized convolutional network for detailed image features. Additionally, Heracles incorporates an attention mechanism in its deeper layers to enhance token interactions, effectively narrowing the performance gap with leading transformers while maintaining lower complexity.
    
    \item \textbf{Parallel Processing:} Heracles employs SSM (Hartley) and convolutional networks concurrently, distinguishing it from traditional Mamba architectures and yielding computational efficiencies.
    
    \item \textbf{Performance Superiority:} Our extensive performance evaluation reveals that Heracles outperforms contemporary transformers, such as LiT, DeIT-B, Biformer, STVit, FDVIT, Wave-VIT, Volo D3, and iFormer, on the ImageNet1K dataset for image classification. This is achieved with significantly reduced computational resources and memory usage. Furthermore, our energy compaction analysis indicates that Heracles can achieve optimal performance with a minimal parameter set. It must be noted that Heracles is the first SSM in the literature to outperform state-of-art transformers, with most SSMs having a performance gap with these transformers.

    \item \textbf{Transfer and Task Learning Excellence:} Heracles exhibits superior performance on classification datasets, such as Flower and Stanford Car, showcasing its efficacy in transfer learning scenarios and validated across diverse task learning domains.

    \item \textbf{SoTA on Time Series:} Heracles beats all state-of-the-art time series models on six standard time series benchmarks, which demonstrates the generalizability of Heracles across all datasets that have spectral properties and have local and global information.

\end{itemize}

\begin{figure*}[htb]%
\centering
\includegraphics[width=0.649\textwidth]{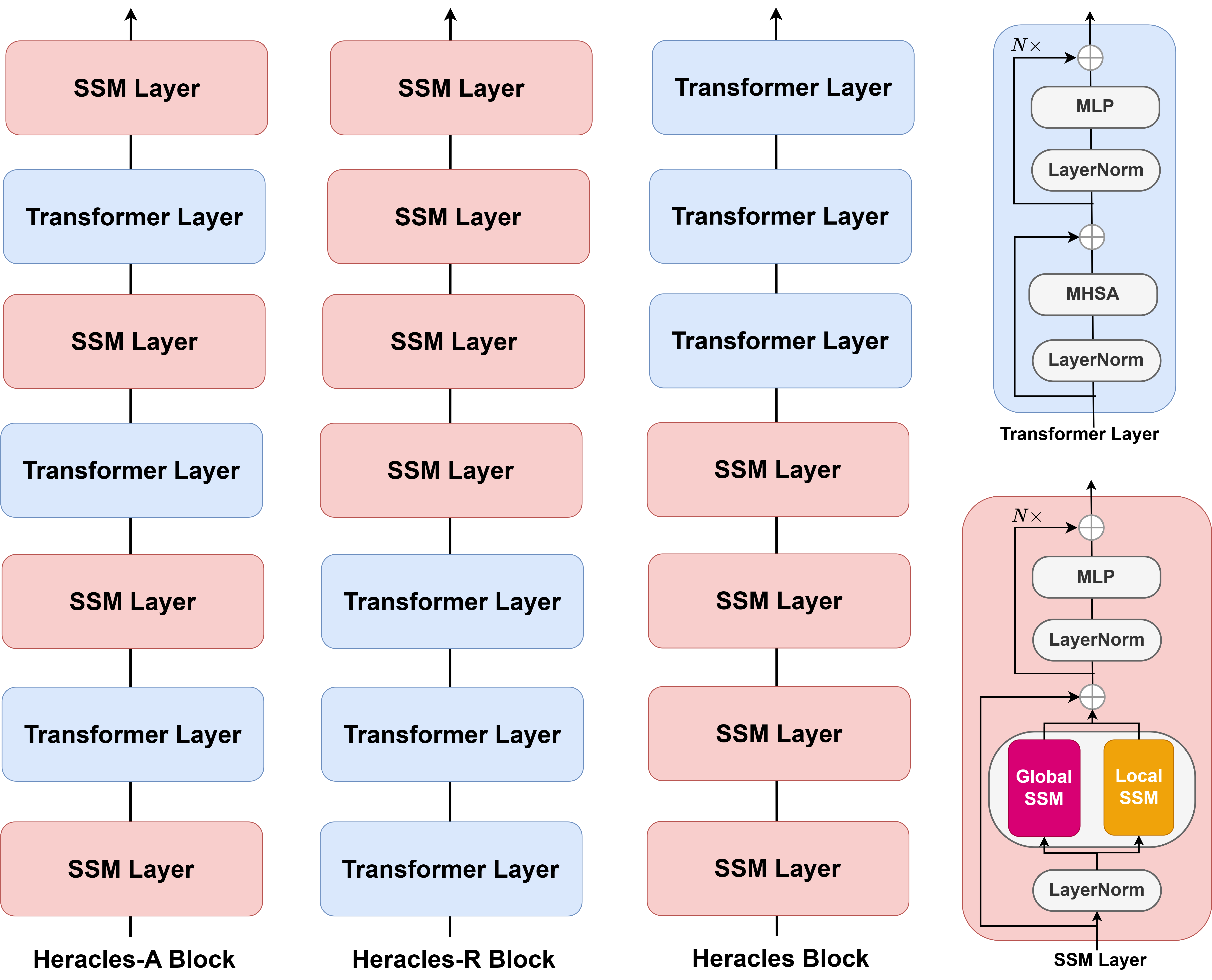}

\vspace{-0.1in}
\caption{This figure illustrates the architectural details of the Heracles model with a SSM Layer and Transformer layer. The SSM Layer comprises a Global SSM using Hartley Transformation to capture global information and a Local SSM using a convolution operator to capture local information. Subsequently, It uses multi-headed attention for message communication between tokens. `A' tends for Alternative, `R' tends for Reverse. }\label{fig:main}
\vspace{-0.5cm}
\end{figure*}
\section{Related Work}

SSMs such as Mamba often encounter instability issues due to their sensitivity to recurrent dynamics.
Vision Mamba\cite{zhu2024vision} and V-Mamba\cite{liu2024vmamba} adapted the Mamba architecture for computer vision tasks, utilizing bi-directional and visual state space models. However, the performance study section reveals a performance gap between Vision Mamba, V-Mamba, and state-of-the-art transformer models like SpectFormer\cite{patro2023spectformer}, SVT\cite{patro2023scattering}, WaveViT\cite{yao2022wave}, and Volo\cite{yuan2022volo}. SiMBA\cite{patro2024simba} addresses this gap by incorporating Mamba for token mixing, replacing attention networks, and leveraging Einstein FFT (EinFFT) for channel mixing. Heracles provides an alternative mechanism to SiMBA for modeling long sequences and bridging performance gaps with state-of-the-art transformers. Heracles uses an approximate Gaussian Kernel as a state space model, whereas SiMBA uses Mamba as the state space model in the architecture. Mamba-based architectures use localized convolution networks in series to the SSM, whereas Heracles uses the convolution in parallel with the SSM, leading to significant efficiencies as demonstrated in the performance studies section. ConvSSM \cite{smith2024convolutional} has been proposed which uses convolutional networks to compute the final kernel function, whereas the instability is still present while computing the kernel functions.


MambaFormer~\cite{park2024can} combines Mamba with attention networks in each layer, whereas Heracles uses SSMs in the initial layers and attention only in the deeper layers, which helps in handling token interactions. Jamba\cite{lieber2024jamba} interleaves transformers, Mamba, and Mixture of Expert (MoE) Mamba layers in the architecture and shows strong performance in NLP tasks. Heracles is a hybrid SSM similar to MambaFormer and Jamba while being scaled to large networks and has been proven as the best-performing SSM for both computer vision and time series data. MambaFormer~\cite{park2024can} retains the instability problem of Mamba and cannot be trained on ImageNet data.



When comparing Heracles with attention-based transformers like ViT\cite{dosovitskiy2020image}, DeIT\cite{touvron2021training}, Swin\cite{liu2021swin}, CSwin\cite{dong2022cswin}, and CVT\cite{wu2021cvt}, it becomes evident that ViT, although pioneering attention-based transformers in computer vision, faced challenges due to its quadratic attention complexity ($O(N^2)$). Subsequent models like DeIT\cite{touvron2021training}, Swin\cite{liu2021swin}, CVT\cite{wu2021cvt}, and CSwin\cite{dong2022cswin} aimed at refining transformer performance in computer vision and specific NLP tasks. Alternatively, MLP Mixer architectures such as MLP Mixers\cite{tolstikhin2021mlp} and DynamicMLP\cite{wang2022dynamixer} sought to replace attention networks with MLP to address these challenges. Recent transformer architectures such as  Flatten Transformer \cite{han2023flatten}, BeIT \cite{bao2021beit}, MixMAE\cite{liu2023mixmae}, SVT \cite{patro2023scattering}, Biformer \cite{zhu2023biformer}, STVit \cite{huang2023vision}, and FDVIT \cite{xu2023fdvit} deliver state-of-the-art performance on several tasks and domains. SpectFormer\cite{patro2023spectformer} and SVT \cite{patro2023scattering} use a mixed variant of transformers by combining frequency and attention layers to achieve state-of-the-art performance. We compare the performance of Heracles with SpectFormer and SVT in the performance studies section.

\section{Method }

\subsection{State Space Modeling:}\label{sec:ss-continuous}
The state space model\cite{gu2021efficiently,gu2023mamba} is commonly known as a linear time-invariant system that maps the input stimulation $x(t) \in \mathcal{R}^L $ to a response $y(t) \ $ through a hidden space $h(t) \in \mathcal{R}^N $. S4\cite{gu2021efficiently} are a recent class of sequence models for deep learning that are broadly related to RNNs, CNNs, and classical state space models. Mathematically,  the Continuous-time Latent State spaces can be modelled as a linear ordinary differential equation that uses an evolution parameter $A \in \mathcal{R}^{N\times N} $ and projection parameters $B \in \mathcal{R}^{N\times 1} $ and $C \in \mathcal{R}^{N\times 1} $ as follows:
  \begin{align} \label{eq:1}
    x'(t) &= \bm{A}x(t) + \bm{B}u(t) \\
    y(t) &= \bm{C}x(t) + \bm{D}u(t) 
  \end{align}
\textbf{Discrete-time SSM:}\label{sec:ss-recurrent}
The discrete form of SSM uses a time-scale parameter $\dt$ to transform continuous parameters A, B, and C to discrete parameters $\Bar{A}, \Bar{B}$ and $ \Bar{C} $ using fixed formula  $\Bar{A}=f_{A}(\dt,A), \Bar{B}=f_{B}(\dt,A,B)$. The pair $f_{A},f_{B}$ is the discretization rule that uses a zero-order hold (ZOH)  for this transformation. The equations are as follows.

\begin{equation}
  \label{eq:2}
\begin{aligned}
  x_{k} &= \bm{\overline{A}} x_{k-1} + \bm{\overline{B}} u_k &
  \bm{\overline{A}} &= (\bm{I} - \dt/2 \cdot \bm{A})^{-1}(\bm{I} + \dt/2 \cdot \bm{A}) &
  \\
  y_k &= \bm{\overline{C}} x_k &
  \bm{\overline{B}} &= (\bm{I} - \dt/2 \cdot \bm{A})^{-1} \dt \bm{B} &
  \bm{\overline{C}} &= \bm{C}
  .
\end{aligned}
\end{equation},
\textbf{where  $I \in \mathcal{R}^{N\times N} $  is the identity matrix}

\textbf{Convolutional Kernel Representation}
\label{sec:ss-convolution}
The discretized form of recurrent SSM in the equation- of {eq:2} is not practically trainable due to its sequential nature. We model continuous convolution as discrete convolution, a linear time-invariant system, to get an efficient representation. 
For simplicity, let the initial state be \( x_{-1} = 0 \).
Then unrolling \eqref{eq:2} explicitly yields: %
\begin{align*}
  x_0 &= \bm{\overline{B}} u_0 &
  x_1 &= \bm{\overline{A}} \bm{\overline{B}} u_0 + \bm{\overline{B}} u_1 &
  x_2 &= \bm{\overline{A}}^2 \bm{\overline{B}} u_0 + \bm{\overline{A}} \bm{\overline{B}} u_1 + \bm{\overline{B}} u_2 & \dots
  \\
  y_0 &= \bm{\overline{C}} \bm{\overline{B}} u_0 &
  y_1 &= \bm{\overline{C}} \bm{\overline{A}} \bm{\overline{B}} u_0 + \bm{\overline{C}} \bm{\overline{B}} u_1 &
  y_2 &= \bm{\overline{C}} \bm{\overline{A}}^2 \bm{\overline{B}} u_0 + \bm{\overline{C}} \bm{\overline{A}} \bm{\overline{B}} u_1 + \bm{\overline{C}} \bm{\overline{B}} u_2
  & \dots
\end{align*}
This can be vectorized into a convolution \eqref{eq:convolution} with an explicit formula for the convolution kernel \eqref{eq:convolution}.
\begin{equation}
  \label{eq:convolution}
  \begin{split}
    y_k &= \bm{\overline{C}} \bm{\overline{A}}^k \bm{\overline{B}} u_0 + \bm{\overline{C}} \bm{\overline{A}}^{k-1} \bm{\overline{B}} u_1 + \dots + \bm{\overline{C}} \bm{\overline{A}} \bm{\overline{B}} u_{k-1} + \bm{\overline{C}}\bm{\overline{B}} u_k
    \\
    y &= \bm{\overline{K}} \ast u = (\bm{\overline{C}}\bm{\overline{B}}, \bm{\overline{C}}\bm{\overline{A}}\bm{\overline{B}}, \dots, \bm{\overline{C}}\bm{\overline{A}}^{L-1}\bm{\overline{B}}).
  \end{split}
\end{equation}
\normalsize
 \( \bm{\overline{K}} \) in equation \eqref{eq:convolution} can be represented as a single (non-circular) convolution which can be computed very efficiently with FFTs. However, computing \( \bm{\overline{K}} \) in \eqref{eq:convolution} is non-trivial and is modelled as an 
 \textbf{SSM convolution kernel} or filter.

The above equation uses different forms of matrices A, B and C, which can be further simplified with a single matrix using FFT. Further, the complex form of FFT can be simplified as a real-valued transform - Heracles uses the Hartley transform for this purpose. Heracles also uses another real-valued transform DCT instead of the Hartley transform, with experimental comparisons captured in the performance studies section. This effectively captures global information in the image, while Heracles also includes another SSM, which uses the spatial convolutional network to capture local information.

\subsection{Global SSM with Hartley Kernel}

The core of Heracles is its SSM, which comprises local and global SSMs. Heracles is strategically designed to capture both global and local features efficiently from the input data. It starts with a Hartley Transform ($\mathcal{H}$) to obtain global information through real frequency components. Simultaneously, the convolutional operator processes the second part of the SSM layer, capturing local image properties. The architecture of Heracles is illustrated in the diagram \ref{fig:intro} and \ref{fig:main}. 
These two data streams are subsequently merged to form a holistic feature representation, which is normalized and passed through a Multi-Layer Perceptron (MLP) layer. The MLP layer generates a new representation of each patch using nonlinear functions. The mathematical formulation of these components is given below.

\textbf{Hartley Operator Module:}
The core part of the Hartley Operator module lies in its Hartley ~\cite{hartley1942more} Transform, which uses the cosine-and-sine (cas) or Hartley kernel to transform the signal from a spatial domain to the Hartley spectral domain or frequency domain. The mathematical form of the Hartley transformer is given as:
\begin{align*}
\label{eq:hartley}
    \mathcal{H}(\omega) &= \frac{1}{2 \pi} \int_ {-\infty}^ {\infty} f(x) cas(\omega x) dx , cas(x) &=sin(x) + cos(x) 
&= \sqrt(2)sin(x+\frac{\pi}{4})  = \sqrt(2)cos(x-\frac{\pi}{4})
\end{align*}
where  $\omega$ is the angular frequency. 
We formulate Heracles using Hartley ~\cite{hartley1942more} Transform as a special case of the Fourier neural operators \cite{li2020fourier}, which transforms real inputs to real outputs, with no intrinsic involvement of complex numbers like the Fourier transform.  Heracles uses a Hartley Neural Operator to learn a mapping between two infinite dimensional spaces from a finite collection of observed input-output pairs  \(\{a_j, u_j\}_{j=1}^N\) using  a function $\mathcal{A}$ to function $\mathcal{U}$
\begin{equation*}
\label{eq:approxmap}
\mathcal{F} : \mathcal{A} \times \Psi \to \mathcal{U}
\quad
\text{equivalently,}
\quad
\mathcal{F}_{\psi} :  \mathcal{A} \to  \mathcal{U}, \quad \psi \in \Psi
\end{equation*}
The input \(a \in \mathcal{A}\) is projected to a higher dimensional representation $v_0(x) = Z(a(x))$ by the local transformation \(Z\) which is usually parameterized by a shallow fully-connected neural network as shown in Figure-~\ref{fig:intro}. Then, we apply several iterations of updates $v_t \mapsto v_{t+1}$. The output $u(x) = Z(v_T(x))$ is the projection of $v_T$ by the local transformation $Z: \mathcal{R}^{d_v} \to \mathcal{R}^{d_u}$. In each iteration, the update $v_t \mapsto v_{t+1}$ is defined as the composition of a non-local integral operator $\mathcal{K}$ and a local, nonlinear activation function $\sigma$. The iterative update to the representation $v_t \mapsto v_{t+1}$ by
\begin{equation}\label{def:int}
v_{t+1}(x) = \sigma \Big( \bigl(\mathcal{K}(a;\psi)v_t\bigr)(x) \Big), \qquad \forall x \in D
\end{equation}

The Kernel integral operator $\mathcal{K}$ is defined as 
\begin{equation}
\label{def:K_int}
\bigl(\mathcal{K}(a;\psi)v_t\bigr)(x) :=  
\int_{D} \kappa\big(x,y,a(x),a(y);\psi \big) v_t(y) \mathrm{d}y, 
\end{equation}
Where $\kappa_{H}$ is a neural network parameterized by $H \in \Psi_{\mathcal{K}}$. Here $\kappa_H$ plays the role of a kernel function which learns from data. Now, the kernel integral operator is defined in Hartley spectral space. 
Let \(\mathcal{H}\) denote the Hartley transform \cite{hartley1942more} of a function \(f: D \to \mathbb{R}^{d_v}\). It has a unique property: its inverse, \(\mathcal{H}^{-1}\), is an involution, meaning \(\mathcal{H}^{-1}\) is its own inverse.
\begin{align*}
    \mathcal{H}_j(k) = \int_{D} f_j(x) cas(\omega x) \mathrm{d}x \text{and}  \quad \mathcal{H}^{-1}_j(x)=\mathcal{H}(\mathcal{H}_j(k) ) 
\end{align*}
For the special case of Green's kernel $\mathcal{K}(s, t) = \mathcal{K}(s-t)$, the integral leads to global convolution. By applying $\kappa_{H}(x,y,a(x),a(y)) = \kappa_{H}(x-y)$ in (\ref{def:K_int}) and applying the convolution theorem, we find that
\[\bigl(\mathcal{K}(a;\psi)v_t\bigr)(x) = \mathcal{H}^{-1} \bigl( \mathcal{H}(\kappa_H) \cdot \mathcal{H}(v_t) \bigr )(x), \qquad \forall x \in D. \]


The Hartley integral operator $\mathcal{K}$ is defined as
\begin{equation}
\label{eq:hio}
\bigl(\mathcal{K}(\psi)v_t\bigr)(x)=   
\mathcal{H}^{-1}\Bigl(R_H \cdot \mathcal{H} (v_t) \Bigr)(x) \quad \forall x \in D 
\end{equation}
where $R_H$ is the real value Hartley transform of the function $\kappa: \bar{D} \to \mathcal{R}^{d_v \times d_v}$ parameterized by \(H \in \Psi_\mathcal{K}\).

\subsubsection{Local SSM with Convolutional Network}
Convolutional Operator module uses convolution neural operators with discrete kernels \cite{raonic2023convolutional} 
$ K_c = \sum_{i,j=1}^{k} k_{ij} \delta_{z_{ij}} $
defined on an $s \times s$ uniform grid with $z_{ij}$ being the resulting grid points and $\delta_x$ is the Dirac measure at point $x$. As explained in ~\cite{raonic2023convolutional}, we can define the convolution operator for a single channel as

\begin{equation}\label{def:conv}
\mathcal{C} =K_c f(x) = (K_c * f)x = \sum_{i,j=1}^{k} k_{ij} f(x - z_{ij})
\end{equation}



\subsubsection {Heracles Block}
The SSM layer of the Heracles combines the global SSM using the Hartley integral operator as mentioned in eq-\ref{eq:hio} where the integral leads to convolution with the Convolutional Operator in eq-\ref{def:conv} to obtain global and local features of the input image. The Heracles Block seamlessly integrates two fundamental operators: the convolutional representation denoted by $\mathcal{C} = K_c f(x)$, defined as per equation (\ref{def:conv}), and the SSM representation $\mathcal{V}$, defined as per equation (\ref{def:int}). This integration is formulated through the following expression:
\noindent
\begin{equation}\label{def:final}
\mathcal{V}_{t+1}(x) = \mathcal{C}_{t} \circ \sigma \Big( \bigl(\mathcal{K}(a;\psi)v_t\bigr)(x) \Big), \qquad \forall x \in D
\end{equation}

\begin{table*}[!tb]



%
%

%
%

\scriptsize
\centering
\caption{The table shows the performance of various vision backbones on the ImageNet1K\cite{deng2009imagenet} dataset for image recognition tasks. $\star$ indicates additionally trained with the Token Labeling objective using MixToken and a convolutional stem ~\cite{wang2022scaled} for patch encoding. We have grouped the vision models into three categories based on their GFLOPs (Small, Base, and Large). The GFLOP ranges: Small (GFLOPs$<$6), Base (6$\leq$GFLOPs$<$9), and Large (10$\leq$GFLOPs$<$30).
}

\setlength{\tabcolsep}{4.5pt}
\begin{tabular}{l|c|c|cc|l|c|c|cc}
\Xhline{2\arrayrulewidth}

Method          & Params & FLOPs & Top-1 & Top-5 & Method          & Params & GFLOPs & Top-1 & Top-5 \\
          & (M) & (G) & (\%) & (\%) &           & (M) & (G) & (\%) & (\%)\\

\hline
\multicolumn{5}{c|} {Small} & \multicolumn{5}{c} {Large} \\ \hline

ResNet-50  \cite{he2016deep}                 & 25.5 & 4.1 & 78.3 & 94.3 &
ResNet-152 \cite{he2016deep}                 & 60.2 & 11.6 & 81.3 & 95.5  \\

BoTNet-S1-50 \cite{srinivas2021bottleneck}   & 20.8 & 4.3 & 80.4 & 95.0 &
ResNeXt101 \cite{xie2017aggregated}    & 83.5 & 15.6 & 81.5 & -     \\
Cross-ViT-S~\cite{chen2021crossViT}& 26.7& 5.6 & 81.0 &- &
gMLP-B~\cite{liu2021pay}                & 73.0& 15.8&81.6&-\\

Swin-T  \cite{liu2021swin}                   & 29.0 & 4.5 & 81.2 & 95.5 & 
DeiT-B \cite{touvron2021training}            & 86.6 & 17.6 & 81.8 & 95.6  \\

ConViT-S \cite{d2021convit}                  & 27.8 & 5.4 & 81.3 & 95.7 &
SE-ResNet-152 \cite{hu2018squeeze}           & 66.8 & 11.6 & 82.2 & 95.9  \\

T2T-ViT-14 \cite{yuan2021tokens}             & 21.5 & 4.8 & 81.5 & 95.7 &
Cross-ViT-B~\cite{chen2021crossViT}          & 104.7& 21.2 & 82.2  &- \\

RegionViT-Ti+ \cite{chen2022regionvit}       & 14.3 & 2.7 & 81.5 & -    & 
ResNeSt-101 \cite{zhang2022resnest}          & 48.3 & 10.2 & 82.3 & -     \\

SE-CoTNetD-50  \cite{li2022contextual}       & 23.1 & 4.1 & 81.6 & 95.8    &
ConViT-B \cite{d2021convit}                  & 86.5 & 16.8 & 82.4 & 95.9  \\

Twins-SVT-S \cite{chu2021twins}              & 24.1 & 2.9 & 81.7 & 95.6 &
PoolFormer-M48 ~\cite{yu2022metaformer}      & 73.0&  11.8&  82.5&-\\

CoaT-Lite Small \cite{xu2021co}              & 20.0 & 4.0 & 81.9 & 95.5 &
T2T-ViTt-24 \cite{yuan2021tokens}            & 64.1 & 15.0 & 82.6 & 95.9  \\

FDViT-S \cite{xu2023fdvit}&	21.5	&2.8&	81.5&	-&
FDViT-B \cite{xu2023fdvit}	&67.8	&11.9	&82.4	&-\\

PVTv2-B2 \cite{wang2022pvt}                & 25.4 & 4.0 & 82.0 & 96.0 & 

TNT-B  \cite{han2021transformer}             & 65.6 & 14.1 & 82.9 & 96.3  \\

LITv2-S~\cite{panfast}                      &28.0 &3.7& 82.0&-  & 
CycleMLP-B4~\cite{chencyclemlp}             &52.0& 10.1& 83.0&- 
\\

MViTv2-T~\cite{li2022mvitv2}                &24.0& 4.7& 82.3&-  &
DeepViT-L  \cite{zhou2021deepvit}           & 58.9 & 12.8 & 83.1 & -     \\

Wave-ViT-S~\cite{yao2022wave}               & 19.8 & 4.3 & 82.7 & 96.2 & 
RegionViT-B \cite{chen2022regionvit}        & 72.7 & 13.0 & 83.2 & 96.1  \\

CSwin-T \cite{dong2022cswin}             & 23.0 & 4.3 & 82.7 & - &
CycleMLP-B5~\cite{chencyclemlp}             & 76.0& 12.3& 83.2&-  \\

DaViT-Ti  ~\cite{ding2022daViT}         & 28.3 & 4.5 &  82.8&- &
ViP-Large/7 ~\cite{hou2022vision}       & 88.0& 24.4& 83.2&- \\

FLatten-CSwin-T \cite{han2023flatten}	&21.0	&4.3&	83.1&	- &
CaiT-S36 \cite{touvron2021going}             & 68.4  & 13.9 & 83.3 & -     \\

 iFormer-S\cite{si2022inception} & 20.0 & 4.8 & 83.4 & 96.6 &
AS-MLP-B ~\cite{lianmlp} & 88.0& 15.2& 83.3&-\\

CMT-S~\cite{guo2022cmt}                     & 25.1 &  4.0& 83.5 &- &
BoTNet-S1-128 \cite{srinivas2021bottleneck}  & 75.1 & 19.3 & 83.5 & 96.5  \\ 

 MaxViT-T~\cite{tu2022maxvit}           & 31.0&  5.6&  83.6&- &
 Swin-B  \cite{liu2021swin}                   & 88.0 & 15.4 & 83.5 & 96.5  \\

 Wave-ViT-S$^\star$~\cite{yao2022wave}       & 22.7 & 4.7 & 83.9 & 96.6 & 
 Wave-MLP-B~\cite{tang2022image}& 63.0& 10.2& 83.6&-  \\

BiFormer-S* \cite{zhu2023biformer}	&26	&4.5	&84.3&	-&
LITv2-B ~\cite{panfast} & 87.0 &13.2 & 83.6&- \\

\cellcolor{gray!15}\textbf{Heracles-C-S$^\star$} & \cellcolor{gray!15}\textbf{21.7} & \cellcolor{gray!15}\textbf{4.1}  & \cellcolor{gray!15}\textbf{84.5} & \cellcolor{gray!15}\textbf{97.0} &

 PVTv2-B4 \cite{wang2022pvt}                & 62.6 & 10.1 & 83.6 & 96.7  \\
 \cline{1-5}
\multicolumn{5}{c|} {Base}& 

ViL-Base ~\cite{zhang2021multi}             & 55.7&  13.4 &83.7& -\\
\cline{1-5}

ResNet-101 \cite{he2016deep}                 & 44.6 & 7.9 & 80.0 & 95.0 &
 Twins-SVT-L \cite{chu2021twins}              & 99.3 & 15.1 & 83.7 & 96.5  \\

BoTNet-S1-59 \cite{srinivas2021bottleneck}   & 33.5 & 7.3 & 81.7 & 95.8 &
Hire-MLP-Large~\cite{Guo_2022_CVPR}         & 96.0& 13.4& 83.8&-  \\

T2T-ViT-19 \cite{yuan2021tokens}             & 39.2 & 8.5 & 81.9 & 95.7 &
RegionViT-B+ \cite{chen2022regionvit}        & 73.8 & 13.6 & 83.8 & -     \\

CvT-21~\cite{wu2021cvt}                     & 32.0 & 7.1 & 82.5 & -    & 

Focal-Base \cite{yang2021focal}              & 89.8 & 16.0 & 83.8 & 96.5  \\

GFNet-H-B~\cite{rao2021global}             &54.0& 8.6 &82.9& 96.2&

PVTv2-B5 \cite{wang2022pvt}                & 82.0 & 11.8 & 83.8 & 96.6  \\

Swin-S  \cite{liu2021swin}                   & 50.0 & 8.7 & 83.2 & 96.2 &
SE-CoTNetD-152 \cite{li2022contextual} & 55.8 & 17.0 & 84.0 & 97.0 \\

Twins-SVT-B \cite{chu2021twins}              & 56.1 & 8.6 & 83.2 & 96.3 &

 DAT-B  ~\cite{xia2022vision} &  88.0& 15.8&  84.0&-\\

SE-CoTNetD-101 \cite{li2022contextual}     & 40.9 & 8.5   & 83.2 & 96.5 &
LV-ViT-M$^\star$ \cite{jiang2021all}         & 55.8 & 16.0 & 84.1 & 96.7  \\

PVTv2-B3 \cite{wang2022pvt}                & 45.2 & 6.9 & 83.2 & 96.5 &
 CSwin-B ~\cite{dong2022cswin} & 78.0  &15.0  & 84.2&- \\

LITv2-M~\cite{panfast} & 49.0& 7.5 & 83.3&-  &
  HorNet-$B_{GF}$~\cite{rao2022hornet}& 88.0& 15.5& 84.3&-  \\

RegionViT-M+ \cite{chen2022regionvit}        & 42.0 & 7.9 & 83.4 & -    & 
DynaMixer-L~\cite{wang2022dynamixer}        & 97.0& 27.4& 84.3&- \\

MViTv2-S~\cite{li2022mvitv2} &35.0 & 7.0 & 83.6&- &
MViTv2-B~\cite{li2022mvitv2} &52.0& 10.2& 84.4&- \\

CSwin-S ~\cite{dong2022cswin}& 35.0 & 6.9  & 83.6&- &
FLatten-CSwin-B \cite{han2023flatten}	&75	&15.0	&84.5	&-\\

DaViT-S ~\cite{ding2022daViT}& 49.7 & 8.8 & 84.2&- &
DaViT-B~\cite{ding2022daViT}& 87.9 & 15.5  &84.6&- \\

VOLO-D1$^\star$  \cite{yuan2022volo}         & 26.6 & 6.8 & 84.2 & - &
CMT-L ~\cite{guo2022cmt}&74.7&19.5& 84.8&-\\

CMT-B ~\cite{guo2022cmt}&45.7&9.3& 84.5&-&
MaxViT-B ~\cite{tu2022maxvit}& 120.0& 23.4&85.0&- \\

FLatten-CSwin-S \cite{han2023flatten}	&35	&6.9	&83.8	&- &
MixMAE \cite{liu2023mixmae}	&88	&16.3	&85.1	&-\\

STViT-S \cite{huang2023vision}	&25	&4.4&	83.6	&-&
STViT-L \cite{huang2023vision}	&95	&15.6	&85.3	&-\\

SiMBA-L(EinFFT)  ~\cite{patro2024simba} & 36.6 & 9.6 & 84.4&- &
VOLO-D2$^\star$ \cite{yuan2022volo}          & 58.7 & 14.1 & 85.2 & -     \\

MaxViT-S~\cite{tu2022maxvit}& 69.0& 11.7& 84.5&- &
BiFormer-B* \cite{zhu2023biformer}	&58	&9.8	&85.4	&-\\

iFormer-B\cite{si2022inception} & 48.0 & 9.4 & 84.6 & 97.0 &
VOLO-D3$^\star$ \cite{yuan2022volo}          & 86.3 & 20.6 & 85.4 & -     \\

Wave-ViT-B$^\star$  \cite{yao2022wave}& 33.5 & 7.2 & 84.8 & 97.1 &
Wave-ViT-L$^\star$ \cite{yao2022wave}  & 57.5 & 14.8 & 85.5 & 97.3  \\

\rowcolor{gray!15}\textbf{Heracles-C-B$^\star$} & \textbf{32.5} & \textbf{6.5}& \textbf{85.2} & \textbf{97.3}
& \textbf{Heracles-C-L$^\star$} & \textbf{54.1} & \textbf{13.4} & \textbf{85.9} & \textbf{97.6} \\
\Xhline{2\arrayrulewidth}
\end{tabular}
\label{tab:imagenet1k_sota}
\vspace{-0.23in}
\end{table*}

\section{Experiments}
 We conducted a comprehensive evaluation of Heracles on key computer vision tasks, including image recognition and instance segmentation. Our assessments for the Heracles model on standard datasets involved: a) Training and evaluating ImageNet1K~\cite{deng2009imagenet} from scratch for the image recognition task. b) Ablation studies to optimize $\alpha$ values and compare different spectral networks within Heracles. c) Ablation analysis comparing the efficacy of initial spectral layers against initial attention or convolution layers. d) Analyze the convolution operator's efficiency and latency regarding the proposed Heracles model. e) Transfer learning on CIFAR-10~\cite{krizhevsky2009learning}, CIFAR-100~\cite{krizhevsky2009learning}, Stanford Cars~\cite{krause20133d}, and Flowers-102~\cite{nilsback2008automated} for image recognition.
f) Finetuning Heracles for downstream instance segmentation tasks. g) Heracles have also been evaluated on six-time series datasets. All experiments were conducted on a hierarchical architecture, currently a state-of-the-art model, with an image size of $224 \times 224 \times 3$.  

\begin{table*}[htb]

\begin{minipage}[t]{.4980492345\textwidth} %

\centering
\scriptsize
\caption{\textbf{SSM SOTA on ImageNet-1K} This table shows the performance of various SSM models for Image Recognition tasks on the ImageNet1K\cite{deng2009imagenet} dataset. We have grouped the vision models into three categories based on their GFLOPs (Small, Base, and Large). The GFLOP ranges: Small (GFLOPs$<$5), Base (5$\leq$GFLOPs$<$10), and Large (10$\leq$GFLOPs$<$30).}
\label{tab:imagenet_sota_ssm}
\begin{tabular}{c|cc|c}
\toprule
Method &  \#Param. & FLOPs  & Top-1 acc. \\

\toprule

HyenaViT-B~\cite{poli2023hyena}   &88M &- & 78.5 \\
S4ND-ViT-B~\cite{nguyen2022s4nd} & 89M & -  & 80.4 \\

TNN-T\cite{qin2022toeplitz} &   6.4M & -  & 72.29\\
TNN-S\cite{qin2022toeplitz} &  23.4M & - & 79.20\\

Vim-Ti\cite{zhu2024vision}  & 7M&- & 76.1 \\ 
Vim-S\cite{zhu2024vision} & 26M &-& 80.5 \\

HGRN-T\cite{qin2024hierarchically} &   6.1M & - &  74.40 \\
HGRN-S\cite{qin2024hierarchically} &  23.7M & - &80.09 \\

PlainMamba-L1 ~\cite{yang2024plainmamba} &  7M &3.0G& 77.9 \\
PlainMamba-L2 ~\cite{yang2024plainmamba} &  25M& 8.1G &81.6\\
PlainMamba-L3 ~\cite{yang2024plainmamba} &   50M& 14.4G& 82.3\\

Mamba-2D-S  ~\cite{li2024mamba}&  24M&- & 81.7\\
Mamba-2D-B ~\cite{li2024mamba} &   92M&- &  83.0\\

VMamba-T\cite{liu2024vmamba} & 22M & 5.6G  & 82.2 \\
VMamba-S\cite{liu2024vmamba} & 44M & 11.2G  & 83.5 \\
VMamba-B\cite{liu2024vmamba} &  75M & 18.0G & 83.2 \\

\hline

LocalVMamba-T ~\cite{huang2024localmamba}  &  26M & 5.7G& 82.7\\
LocalVMamba-S ~\cite{huang2024localmamba} &  50M & 11.4G& 83.7\\


SiMBA-S(EinFFT)  ~\cite{patro2024simba} &  15.3M &2.4G  & 81.7 \\
SiMBA-B(EinFFT)  ~\cite{patro2024simba} &  22.8M &5.2G  & 83.5 \\
SiMBA-L(EinFFT)  ~\cite{patro2024simba} & 36.6M & 9.6G & 84.4 \\

Heracles-Hartley-S(Ours) & 21.7 & 4.1 & \textbf{84.4} \\ 
Heracles-Hartley-B(Ours) & 32.5 & 6.5 & \textbf{85.3}\\
Heracles-Hartley-L (Ours) &54.1 & 13.4 &\textbf{85.7}\\
\rowcolor{gray!15}Heracles-Cosine-S(Ours) & 21.7 & 4.1 & \textbf{84.5} \\ 
\rowcolor{gray!15}Heracles-Cosine-B(Ours) & 32.5 & 6.5 & \textbf{85.2}\\
\rowcolor{gray!15} Heracles-Cosine-L(Ours) &54.1 & 13.4 &\textbf{85.9}\\
\bottomrule
\end{tabular}
\end{minipage}
\begin{minipage}[t]{0.5089492345\textwidth}

\begin{minipage}[t]{0.989492345\textwidth}
\scriptsize
\centering
\caption{Larger Scale Evaluation}\label{tab:large_scale}
\setlength{\tabcolsep}{3.0pt}
\begin{tabular}{lcccc}
\hline
Model &  \tabincell{c}{Params \\ (M)} &  \tabincell{c}{FLOPs \\ (G)}  &  \tabincell{c}{Top-1 \\ (\%)}    \\
\hline
Lit-22B$^\dagger$\cite{zhai2022lit} & 22000 & - & 85.9  \\
\rowcolor{gray!15} Heracles-Cosine-H$^\star$(Ours) &156.7 & 39.3 &\textbf{86.4}\\
ViT-22B$^\dagger$\cite{dehghani2023scaling} & 21743 &-  & 89.5\\ \hline
\end{tabular}
\end{minipage}


\begin{minipage}[t]{0.989492345\textwidth}
\scriptsize
\centering
  \caption{\textbf{Results on transfer learning datasets}. We report the top-1 accuracy on the four datasets. }\label{tab:transfer_learning}%
    \setlength{\tabcolsep}{2.5pt}
    \begin{tabular}{c| cccc}
    \toprule
        Model  &  {CIFAR 10}   & {CIFAR 100} & {Flowers 102} & {Cars 196} \\
    \midrule
    ResNet50~\cite{he2016deep}    & - & - & 96.2  & 90.0 \\
    ViT-B/16~\cite{dosovitskiy2020image}  & 98.1  & 87.1  & 89.5  & - \\
    ViT-L/16~\cite{dosovitskiy2020image}       & 97.9  & 86.4  & 89.7  & - \\
    Deit-B/16~\cite{touvron2021training}     & {99.1}  & {90.8}  & 98.4  & 92.1 \\
    ResMLP-24~\cite{touvron2022resmlp}    & 98.7  & 89.5  & 97.9  & 89.5 \\       
     
     GFNet-XS~\cite{rao2021global}   &  98.6  &  89.1  &   98.1  &  92.8 \\
      GFNet-H-B~\cite{rao2021global}   &  99.0  &  90.3  &   98.8  & 93.2 \\\midrule
     Heracles-C-B   &  \textbf{99.2}  &  \textbf{91.1}  &   \textbf{98.9}  &  \textbf{93.5} \\
     \bottomrule
    \end{tabular}%
    \end{minipage}

\begin{minipage}[t]{0.989492345\textwidth}
\scriptsize 
 \centering
\caption{The performances of various vision backbones on COCO val2017 dataset for the downstream instance segmentation task such as Mask R-CNN 1x \cite{he2017mask} method. We adopt Mask R-CNN as the base model, and the bounding box and mask Average Precision  (\emph{i.e.}, $AP^b$ and $AP^m$)  are reported for evaluation}\label{tab:task_learning}

\setlength{\tabcolsep}{3.0pt}
\begin{tabular}{l|cccccc}
\Xhline{2\arrayrulewidth}
Backbone   &  $AP^b$ & $AP^b_{50}$ & $AP^b_{75}$ & $AP^m$  & $AP^m_{50}$ & $AP^m_{75}$   \\ \hline

ResNet50 \cite{he2016deep}                  & 38.0 & 58.6  & 41.4  & 34.4 & 55.1  & 36.7  \\
Swin-T   \cite{liu2021swin}                  & 42.2 & 64.6  & 46.2  & 39.1 & 61.6  & 42.0 \\
Twins-SVT-S \cite{chu2021twins}               & 43.4 & 66.0  & 47.3  & 40.3 & 63.2  & 43.4\\
LITv2-S~\cite{panfast} & 44.9  &-&-&40.8  &-&- \\
RegionViT-S \cite{chen2022regionvit}          & 44.2 & -     & -     & 40.8 & -     & - \\
PVTv2-B2  \cite{wang2022pvt}               
& 45.3 & 67.1  & 49.6  & 41.2 & 64.2  & 44.4\\ 
\rowcolor{gray!15}Heracles-C-S
& \textbf{45.9}& \textbf{67.8}& \textbf{50.2}& \textbf{41.6} &\textbf{65.0} &\textbf{45.2}\\

\hline


\end{tabular}
    \end{minipage}
    \end{minipage}  
    \vspace{-0.15in}
\end{table*}

\subsection{SOTA Comparison with all kind of Networks on ImageNet 1K}

We conducted performance comparisons on the ImageNet 1K dataset, comprising 1.2 million training images and 50,000 validation images across 1000 categories. Results in Table \ref{tab:imagenet1k_sota} categorize architectures by size: small (<6 GFlops), base (6-10 GFlops), and large (>10 GFlops). In the small category, Heracles-C-Small achieves 84.5\% accuracy, outperforming Wave-ViT-S (83.9\%), Max-ViT-T (83.6\%), and iFormer-S (83.4\%) with lower GFlops and parameters. For the base size, Heracles-C-Base attains 85.2\% accuracy, surpassing Wave-ViT-B (84.8\%), iFormer-B (84.6\%), and Max-ViT-S (84.5\%) while maintaining fewer parameters and GFlops. In the large category, Heracles-C-Large reaches 85.9\% top-1 accuracy, surpassing Volo-D3 (85.4\%), Wave-ViT-L (85.5\%), and Max-ViT-B (85.0\%). Heracles-C-Huge reaches 86.4\% top-1 accuracy, outperforming LIT-22B (85.9\%), whereas ViT-22B has reached 89\% top-1 accuracy, but both require 22 billion parameters, which is close to 100x the number of parameters compared to Heracles. These comparisons demonstrate Heracles-C's superior performance across various architectures, validated on the ImageNet dataset with an image size of 224x224 pixels.

 We also compare different kinds of transformer architectures, such as Convolutional Neural Networks (CNNs), Transformer architectures (attention-based models), MLP Mixers, and Spectral architectures; Heracles-c consistently outperforms its counterparts in the table. We include distillation- and non-distillation-based models and other kinds of models in the comparison. For instance, Heracles-C achieves better top-1 accuracy and parameter efficiency compared to CNN architectures like ResNet 152~\cite{he2016deep}, ResNeXt ~\cite{xie2017aggregated}, and ResNeSt in terms of top-1 accuracy and number of parameters. Among attention-based architectures, MaxViT has been recognized as the best performer, surpassing models like DeiT, Cross-ViT, DeepViT, T2T, etc., with a top-1 accuracy of 85.0. However, Heracles-C achieves an even higher top-1 accuracy of 85.9 with less than half the number of parameters. In the realm of MLP Mixer-based architectures, DynaMixer~\cite{wang2022dynamixer} emerges as the top-performing model, surpassing  MLP-mixer, gMLP, CycleMLP, Hire-MLP, AS-MLP, WaveMLP, PoolFormer and  DynaMixer-L with a top-1 accuracy of 84.3\%. In comparison, Heracles-C-L outperforms DynaMixer with a top-1 accuracy of 85.9\%. We have captured the large scale performance comparison of Heracles with LiT-22B and ViT-22B in table \ref{tab:large_scale} - this shows that only ViT-22B outperforms Heracles, but needs order of magnitude more parameters (156 million vs 22 billion). It must be noted that Heracles beats LiT-22B with only 156 million parameters. 

\subsection{SOTA Comparison with SSMs}

The table-\ref{tab:imagenet_sota_ssm} presents the state-of-the-art (SOTA) performance of various Structured State Space Models (SSMs) on the ImageNet-1K dataset for image classification tasks. These SSM models are grouped based on their computational complexity measured in GFLOPs (Small, Base, and Large). \textbf{references} Notable models include HyenaViT-B\cite{poli2023hyena}, S4ND-ViT-B\cite{nguyen2022s4nd}, Vim-S~\cite{zhu2024vision}, HGRN-S\cite{qin2024hierarchically}, PlainMamba-L3\cite{yang2024plainmamba}, Mamba-2D-B\cite{li2024mamba}, VMamba-S~\cite{liu2024vmamba}, LocalVMamba-S~\cite{huang2024localmamba}, ViM2-B~\cite{behrouz2024mambamixer}, and SiMBA-L(EinFFT), achieving top-1 accuracy ranging from 78.5\% to 84.5\%. We compare hierarchical architectures using Heracles-C-S with small (S), Heracles-C-Base with base (B), Heracles-C-Large, and Heracles-C-Huge computing. We divide the above according to the GFLops computation like GFNet. We also compare the hierarchical or similar-sized architecture of various vision transformers in table-\ref{tab:imagenet_sota_ssm}. For the Base version, we compare Heracles-C-B with GFNet-H-B, LiT-M, Lit v2-M, iFormer-B, WaveVit-B, and Wave-MLP-B. Heracles-C outperforms all the above w.r.t top-1 accuracy with fewer parameters and GFlops. We also show similar comparisons for small and large variants of hierarchical architectures in the table. These models demonstrate competitive performance while varying in parameter counts and computational complexity, showcasing the effectiveness of SSMs for image classification tasks with improved computational efficiency.

\subsection{SOTA on Multivariate Time Series Datasets}
\linespread{1.2}
\begin{table*}[t]
\centering
 \caption{Multivariate long-term forecasting results with Heracles. We use prediction lengths $T\in \{24, 36, 48, 60\}$ for ILI dataset and $T\in \{96, 192, 336, 720\}$ for the others. The best results are in \textbf{bold}, and the second best is \underline{underlined}.}
	\label{tab:MTS_supervised}
	\resizebox{\linewidth}{!}{
		\begin{tabular}{cc|c|cc|cc|cc|cc|cc|cc|cc|cc|cc|ccc}
			\cline{2-23}
			&\multicolumn{2}{c|}{Models}& \multicolumn{2}{c|}{HFormer}& \multicolumn{2}{c|}{TimesNet}& \multicolumn{2}{c|}{Crossformer}& \multicolumn{2}{c|}{PatchTST}& \multicolumn{2}{c|}{ETSFormer}& \multicolumn{2}{c|}{DLinear}& \multicolumn{2}{c|}{FEDFormer}& \multicolumn{2}{c}{Autoformer}&\multicolumn{2}{c}{Pyraformer}&\multicolumn{2}{c}{MTGNN}& \\
			\cline{2-23}
			&\multicolumn{2}{c|}{Metric}&MSE&MAE&MSE&MAE&MSE&MAE&MSE&MAE&MSE&MAE&MSE&MAE&MSE&MAE&MSE&MAE&MSE&MAE&MSE&MAE\\
			\cline{2-23}
			&\multirow{4}*{\rotatebox{90}{ETTm1}}& 96 &\textbf{0.326}&\textbf{0.369} & \underline{0.338} & \underline{0.375} & 0.349 & 0.395 & 0.339 & 0.377 & 0.375 & 0.398 & 0.345 & 0.372 & 0.379 & 0.419 & 0.505 & 0.475 & 0.543 & 0.510 & 0.379 & 0.446 \\
            &\multicolumn{1}{c|}{}& 192 & \underline{0.376}&\underline{0.397} & \textbf{0.374} & \textbf{0.387} & 0.405 & 0.411 & 0.376 & 0.392 & 0.408 & 0.410 & 0.380 & 0.389 & 0.426 & 0.441 & 0.553 & 0.496 & 0.557 & 0.537 & 0.470 & 0.428 \\
            &\multicolumn{1}{c|}{}& 336 & \textbf{0.408} &0.418  & 0.410 & \textbf{0.411} & 0.432 & 0.431 & 0.408 & 0.417 & 0.435 & 0.428 & 0.413 & \underline{0.413} & 0.445 & 0.459 & 0.621 & 0.537 & 0.754 & 0.655 & 0.473 & 0.430 \\
            &\multicolumn{1}{c|}{}& 720 & 0.482 &0.460  & \underline{0.478} & \textbf{0.450} & 0.487 & 0.463 & 0.499 & 0.461 & 0.499 & 0.462 & \textbf{0.474} & \underline{0.453} & 0.543 & 0.490 & 0.671 & 0.561 & 0.908 & 0.724 & 0.553 & 0.479 \\
			\cline{2-23}
			&\multirow{4}*{\rotatebox{90}{ETTm2}}& 96 &\textbf{0.176} & \textbf{0.264}& \underline{0.187}& \underline{0.267}& 0.208& 0.292& 0.192& 0.273& 0.189 & 0.280 & 0.193 & 0.292 & 0.203 & 0.287 & 0.255 & 0.339 & 0.435 & 0.507 & 0.203 & 0.299 \\
            &\multicolumn{1}{c|}{} & 192 &\textbf{0.245} &\underline{0.310} & \underline{0.249}& \textbf{0.309}& 0.263& 0.332& 0.252& 0.314& 0.253 & 0.319 & 0.284 & 0.362 & 0.269 & 0.328 & 0.281 & 0.340 & 0.730 & 0.673 & 0.265 & 0.328 \\ 
            &\multicolumn{1}{c|}{}& 336 &\textbf{0.305} &\textbf{0.348} & \underline{0.321}& \underline{0.351}& 0.337& 0.369& 0.318& 0.357& 0.314 & 0.357 & 0.369 & 0.427 & 0.325 & 0.366 & 0.339 & 0.372 & 1.201 & 0.845 & 0.365 & 0.374 \\
            &\multicolumn{1}{c|}{}&720 & \textbf{0.404}&\underline{0.405} & \underline{0.408}& \textbf{0.403}& 0.429& 0.430& 0.413& 0.416& 0.414 & 0.413 & 0.554 & 0.522 & 0.421 & 0.415 & 0.433 & 0.432 & 3.625 & 1.451 & 0.461 & 0.459 \\
            \cline{2-23}
			&\multirow{4}*{\rotatebox{90}{ETTh1}}& 96 & \textbf{0.377} & \textbf{0.400} & \underline{0.384} & \underline{0.402} & 0.384 & 0.428 & 0.385 & 0.408 & 0.494 & 0.479 & 0.386 & 0.400 & 0.376 & 0.419 & 0.449 & 0.459 & 0.664 & 0.612 & 0.515 & 0.517 \\
			&\multicolumn{1}{c|}{}&192 & \textbf{0.427} &\textbf{0.428}  & \underline{0.436} & \underline{0.429} & 0.438 & 0.452 & 0.431 & 0.432 & 0.538 & 0.504 & 0.437 & 0.432 & 0.420 & 0.448 & 0.500 & 0.482 & 0.790 & 0.681 & 0.553 & 0.522 \\
			&\multicolumn{1}{c|}{}& 336 & \textbf{0.462} & \textbf{0.446} & \underline{0.491} & \underline{0.469} & 0.495 & 0.483 & 0.485 & 0.462 & 0.574 & 0.521 & 0.481 & 0.459 & 0.459 & 0.465 & 0.521 & 0.496 & 0.891 & 0.738 & 0.612 & 0.577 \\
			&\multicolumn{1}{c|}{}& 720 & \textbf{0.475} &\textbf{0.475}  & \underline{0.521} & \underline{0.500} & 0.522 & 0.501 & 0.497 & 0.483 & 0.562 & 0.535 & 0.519 & 0.516 & 0.506 & 0.507 & 0.514 & 0.512 & 0.963 & 0.782 & 0.609 & 0.597 \\
			\cline{2-23}
			&\multirow{4}*{\rotatebox{90}{ETTh2}}& 96 & \textbf{0.291} & \textbf{0.338} & 0.340 & \underline{0.374} & 0.347 & 0.391 & 0.343 & 0.376 & 0.340 & 0.391 & \underline{0.333} & 0.387 & 0.358 & 0.397 & 0.346 & 0.388 & 0.645 & 0.597 & 0.354 & 0.454 \\ 
            &\multicolumn{1}{c|}{} & 192 & \textbf{0.375} & \textbf{0.391} & \underline{0.402} & \underline{0.414} & 0.419 & 0.427 & 0.405 & 0.417 & 0.430 & 0.439 & 0.477 & 0.476 & 0.429 & 0.439 & 0.456 & 0.452 & 0.788 & 0.683 & 0.457 & 0.464 \\
            &\multicolumn{1}{c|}{}& 336 & \textbf{0.377} & \textbf{0.406} & \underline{0.452} & \underline{0.452} & 0.449 & 0.465 & 0.448 & 0.453 & 0.485 & 0.479 & 0.594 & 0.541 & 0.496 & 0.487 & 0.482 & 0.486 & 0.907 & 0.747 & 0.515 & 0.540 \\
            &\multicolumn{1}{c|}{}& 720 &\textbf{0.412}  & \textbf{0.434} & \textbf{0.462} & \underline{0.468} & \underline{0.479} & 0.505 & 0.464 & 0.483 & 0.500 & 0.497 & 0.831 & 0.657 & 0.463 & 0.474 & 0.515 & 0.511 & 0.963 & 0.783 & 0.532 & 0.576 \\
			\cline{2-23}
			&\multirow{4}*{\rotatebox{90}{Electricity}}& 96 & \textbf{0.145} & \textbf{0.240} & 0.168 & 0.272 & 0.185 & 0.288 & \underline{0.159} & \underline{0.268} & 0.187 & 0.304 & 0.197 & 0.282 & 0.193 & 0.308 & 0.201 & 0.317 & 0.386 & 0.449 & 0.217 & 0.318\\
            &\multicolumn{1}{c|}{}& 192 &\textbf{0.159}  & \textbf{0.252} & 0.198 & 0.300 & 0.211 & 0.312 & \underline{0.195} & \underline{0.296} & 0.212 & 0.329 & 0.209 & 0.301 & 0.214 & 0.329 & 0.231 & 0.338 & 0.376 & 0.443 & 0.260 & 0.348 \\
            &\multicolumn{1}{c|}{}& 336 & \textbf{0.173} & \textbf{0.268} & 0.198 & 0.300 & 0.211 & 0.312 & \underline{0.195} & \underline{0.296} & 0.212 & 0.329 & 0.209 & 0.301 & 0.214 & 0.329 & 0.231 & 0.338 & 0.376 & 0.443 & 0.260 & 0.348 \\
            &\multicolumn{1}{c|}{}& 720 & \textbf{0.214} & \textbf{0.305} & 0.220 & 0.320 & 0.223 & 0.335 & \underline{0.215} & \underline{0.317} & 0.233 & 0.345 & 0.245 & 0.333 & 0.246 & 0.355 & 0.254 & 0.361 & 0.376 & 0.445 & 0.290 & 0.369 \\
			\cline{2-23}
   
			&\multirow{4}*{\rotatebox{90}{Weather}}&96 & \textbf{0.171} & \textbf{0.217} & \underline{0.172} & \underline{0.220} & 0.191 & 0.251 & 0.171 & 0.230 & 0.197 & 0.281 & 0.196 & 0.255 & 0.217 & 0.296 & 0.266 & 0.336 & 0.622 & 0.556 & 0.230 & 0.329 \\ 
   
            &\multicolumn{1}{c|}{} & 192 & \textbf{0.218} & \textbf{0.258 } & \underline{0.219} & \underline{0.261} & 0.219 & 0.279 & 0.219 & 0.271 & 0.237 & 0.312 & 0.237 & 0.296 & 0.276 & 0.336 & 0.307 & 0.367 & 0.739 & 0.624 & 0.263 & 0.322 \\
            
            &\multicolumn{1}{c|}{}& 336 & \textbf{0.270} & \textbf{0.296} & \underline{0.280} & \underline{0.306} & 0.287 & 0.332 & 0.277 & 0.321 & 0.298 & 0.353 & 0.283 & 0.335 & 0.339 & 0.380 & 0.359 & 0.395 & 1.004 & 0.753 & 0.354 & 0.396 \\ 
            &\multicolumn{1}{c|}{}&720 & \textbf{0.343} & \textbf{0.344} & 0.365 & 0.359 & 0.368 & 0.378 & 0.365 & 0.367 & 0.352 & 0.288 & \underline{0.345} & \underline{0.381} & 0.403 & 0.428 & 0.419 & 0.428 & 1.420 & 0.934 & 0.409 & 0.371 \\ 
			\cline{2-23}
		\end{tabular}
  }
	
\end{table*}

We conducted a comprehensive evaluation of our State Space model, Heracles, on seven benchmark standard datasets widely used for Multivariate Time Series Forecasting, including Electricity, Weather, Traffic, and four ETT datasets (ETTh1, ETTh2, ETTm1, and ETTm2), as presented in Table \ref{tab:MTS_supervised}. Our evaluation compares Heracles with various state-of-the-art models, including Transformer-based methods like PatchTST \cite{nie2022time}, CrossFormer \cite{zhang2022crossformer}, FEDFormer \cite{zhou2022fedformer}, ETSFormer \cite{woo2022etsformer}, PyraFormer \cite{liu2021pyraformer}, and AutoFormer \cite{chen2021autoformer}. Additionally, CNN-based methods such as TimeNet \cite{wu2022timesnet}, graph-based methods like MTGNN \cite{wu2020connecting}, and MLP-based models like DLinear \cite{zeng2023transformers} are included in the comparison.

Heracles demonstrates superior performance across multiple evaluation metrics, including Mean Squared Error (MSE) and Mean Absolute Error (MAE), outperforming the state-of-the-art models. These results underscore the versatility and effectiveness of the Heracles architecture in handling diverse time series forecasting tasks and modalities, solidifying its position as a leading model in the field. While presenting our results, it's important to note that due to space limitations in Table \ref{tab:MTS_supervised}, we couldn't include some recent methods in the Time series domain, such as FourierGNN \cite{yi2023fouriergnn}, CrossGNN \cite{huang2023crossgnn}, TimeGPT \cite{liao2024timegpt}, TiDE \cite{das2023long}, SciNet \cite{liu2022scinet}, and FreTS \cite{yi2024frequency}. For a fair comparison, we utilized the code from PatchTST \cite{nie2022time}, and the results were based on a lookup window of size 96 for all datasets.

\subsection{Transfer Learning Studies}

 We have used a pre-trained version of Heracles-C, which is trained from scratch on the ImageNet 1K dataset, and we validate the model's performance on various other datasets on the image classification task. We chose CIFAR-10, CIFAR-100, Oxford Flower, and Stanford Car datasets. For CIFAR-10, We compare different models such as ViT, DeIT, and GFNet. We show that Heracles-C outperforms the above on CIFAR-10 with a top-1 accuracy of 99.2. Similarly for the CIFAR-100 dataset, we show that Heracles-C has a top-1 accuracy 91.2. CosinerFormer outperforms the same transformers on both Flower and Car datasets, as shown in table-~\ref{tab:transfer_learning}.

\subsection{Task Learning Studies}
We show the performance of Heracles-C with other transformers on task learning by taking instance segmentation tasks. Here, we use Heracles-C as a pre-trained model and finetune it to the MSCOCO dataset. We use a Mask-RCNN-based model, such as segmentation. We show the bounding box and mask average precision metrics for evaluation. We compare the convolution-based architecture (ResNet) and transformers such as SWIN, TWIN, LiT v2, PVT V2, and vision ViT. We compare Heracles-C with the above transformers in the bounding box and mask average precision metrics, with Heracles-C emerging as the state-of-the-art transformer as shown in table-~\ref{tab:task_learning}.

\section{Conclusion}

In conclusion, the proposed Heracles introduces a novel hybrid approach by combining Discrete Cosine Transformation (DCT)/Hartley transform with convolutional nets in initial layers and integrating self-attention in deeper layers. This architecture allows Heracles to capture global and local information representations and capturing token interactions effectively, addressing a key limitation of traditional transformers with O($n^2$) complexity. We have shown that Heracles has state-of-the-art performance on several datasets, including ImageNet, transfer learning with Stanford Car, Flower, etc., task learning with image segmentation and object detection, and six-time series datasets. We have also established that Heracles is now the best-performing SSM across these datasets. Heracles opens up research in new architectures with several possibilities that could be explored, including discretization of state space models as well as stability of SSMs at large scale network size. 


{\small 
\bibliographystyle{plain}
\bibliography{egbib}
}

\clearpage
\appendix
\section*{Appendix}
\section{Limitations}

Heracles uses the deeper attention layers and consequently, its complexity is comparable to transformers and not that of SSMs. Heracles uses convolutional networks for capturing the local image properties. Alternatives based on frequency domain approaches could be explored for better representational efficiency. 

\section{Performance Analysis}
This document provides a comprehensive analysis of the vanilla transformer architecture and explores various versions. The architecture comparisons are presented in Table-\ref{tab:arch}, shedding light on the differences and capabilities of each version. The document also delves into the training configurations, encompassing transfer learning, task learning, and finetuning tasks. The dataset information utilized for transformer learning is presented in Table-~\ref{tab:transfer_learning_dataset}, providing insights into dataset sizes and relevance to different applications. Moving to the results section, we showcase the finetuned model outcomes, where models are initially trained on 224 x 224 images and finetuned on 384 x 384 images. The performance evaluation, as depicted in Table-~\ref{tab:finetune}, encompasses accuracy metrics, number of parameters(M), and Floating point operations(G).

\begin{table}[htb]
\scriptsize
\centering
\caption{Ablation Analysis on Local and Global Architecture: Initial application of cosine Transform, followed by the integration of a convolution operator, and further experimentation with stacked convolution layers to attain optimal results}\label{tab:dct_conv}
\setlength{\tabcolsep}{3.0pt}
\begin{tabular}{lccccc}
\hline
Model &  \tabincell{c}{Params \\ (M)} &  \tabincell{c}{FLOPs \\ (G)}  &  \tabincell{c}{Top-1 \\ (\%)}   &  \tabincell{c}{Top-5 \\ (\%)}  \\
\hline
Heracles-Cosine & 22.04& 4.1& 84.2&96.94 \\
Heracles-Conv & 21.7M & 4.1& 84.0 &95.7\\
Heracles-Cosine-Conv(Series) & 21.9 &4.1& 84.16& 96.84\\
Heracles-Cosine-Conv(Parallel) & 21.71 &4.1& 84.46& 97.01\\
\hline
\end{tabular}
\end{table}

\begin{table}[htb]
\scriptsize
\centering
\caption{This table shows the ablation analysis of various spectral transformations, including real and complex transforms in Heracles architecture such as Fast Fourier Convolution (FFC), Fourier, Wavelet, DTCWT, Cosine, and Hartley transforms for small-size stage networks architecture. This indicates that Heracles-Cosine performs better than other kinds of networks.}\label{tab:spectral_nets}
\setlength{\tabcolsep}{3.0pt}
\begin{tabular}{lccccc}
\hline
Model & {Params(M)} &  {FLOPs(G)} & {Top-1(\%)}   & {Top-5(\%)} \\
\hline
Fourier Net & 21.17 &3.9& 84.02& 96.77\\
DTCWT &{22.0M} & {3.9}  & {84.2} & {96.9}\\
Wavelet & 21.59& 3.9& 83.70&96.56\\
\rowcolor{gray!15}Heracles-Hartley & 21.7M & 4.1 & \textbf{84.4} &\textbf{96.9}\\
\rowcolor{gray!15}Heracles-Cosine & 21.7M & 4.1 & \textbf{84.5} &\textbf{97.0}\\
\hline
\end{tabular}
\end{table}

\begin{table}
\scriptsize
 \centering
\caption{We have shown different values of alpha in hierarchical architecture starting from $\alpha=0$ means it is not spectral all are attention layers then $\alpha=1$ which means the first stage is spectral and the remaining all are attention layers. Similarly for $\alpha=2$, which means the first and second stages are spectral layers and the remaining are attention, and $\alpha=4$, which means all stages are spectral and there are no attention layers.}\label{tab:alpha}
 \vspace{-0.4em}
 \setlength{\tabcolsep}{6.0pt}
\begin{tabular}{lccccc}
\hline
Model & {Params(M)} &  {FLOPs(G)} & {Top-1(\%)}   & {Top-5(\%)} \\
\hline
Heracles-C-S-a0 & 22.22& 3.9& 83.12&96.67 \\
\rowcolor{gray!15}Heracles-C-S-a1 & 21.72 & 4.1 & \textbf{84.46} &\textbf{97.0}\\
Heracles-C-S-a2 & 21.6 &3.9& 84.10& 96.87\\
Heracles-C-S-a3 & 19.41 &3.5& 82.90&96.46 \\
Heracles-C-S-a4 & 17.03& 3.3& 82.45&96.12 \\

\hline
\end{tabular}
\end{table}

\begin{table}
\scriptsize
\centering
\caption{\textbf{Initial Attention vs Cosine vs Convolutional Layer:} This table compares Heracles-C transformer with initial spectral layers and later attention layers, Heracles-C-Inverse with initial attention layers and later spectral layers, and  Heracles-C with initial convolutional layers. Also, we show an alternative spectral layer and attention layer. This shows that the Initial spectral layer works better compared to the rest.  }
\label{tab:init_conv_atten}
\setlength{\tabcolsep}{5.0pt}
\begin{tabular}{lcccc}

\hline

Model & {Params(M)} &  {FLOPs(G)} & {Top-1(\%)}   & {Top-5(\%)} \\
\hline
\rowcolor{gray!15}Heracles-C-S (init-spectral) & 22.0M & 3.9 & 84.2 &96.9\\
Heracles-C-Init-CNN & 21.7M & 4.1& 84.0 &95.7\\
Heracles-C-Inverse & 21.8M & 3.9 & 83.1 &94.6\\
Heracles-C-Alternate & 22.4M & 4.6 & 83.4 &95.0\\


\hline
\end{tabular}  
   \end{table}

\section{Ablation Studies on SSM Layers}
We performed a study on various spectral transforms including real and complex domain such as Fourier transform, Wavelet transform, Dual-Tree Complex Wavelet Transform (DTCWT), Cosine Transform, and Hartley Transform. We start the spectral layer with complex domain like a Fourier transform similar to FNet\cite{lee2021fnet}, Fast fourier Convolution\cite{chi2020fast}, GFNet and AFNO~\cite{guibas2021efficient} which shows a top-1 accuracy of 84.02 on a small-size hierarchical model. We investigate with another complex domain transform such as DTCWT and obtain  top-1 accuracy of 84.2. We also investigate with use the discrete wavelet transform based network and obtain 83.7 top-1 accuracy. We investigate with real transform such as Cosine and Hartley transform. We also show that the Heracles-C  has a top-1 accuracy of 84.5\% and Heracles-H. We conclude thus that the cosine transform learns better to provide a better spectral representation compared to other alternatives discussed above, as shown in table-~\ref{tab:spectral_nets}. 


The Heracles-C architecture is characterized by a staged approach, comprising a total of $\alpha$ initial spectral layers/stages followed by $L - \alpha$ attention layers/stages. Here, $L$ represents the total number of layers/stages in the architecture. The integration of spectral and attention layers within the staged architecture is instrumental in achieving its remarkable performance. Another ablation study is to decide the value of $\alpha$. This is captured in table-~\ref{tab:alpha}, which shows studies with a hierarchical network having different $\alpha$ values which indicate the number of initial spectral layers. $\alpha$ value of 1 for stage-1 with all spectral layers, while $\alpha$=2 with all layers of stage-1 and stage-2 being spectral layers. Similarly, for $\alpha=3$, all layers of stage-1, stage-2, and stage-3 are spectral layers.

\subsection{Local and Global Operators}
We initially employed the spectral network incorporating cosine transformation (Heracles-Cosine) to capture global information and  the convolutional network (Heracles-Conv)  to capture local information as shown in table-\ref{tab:dct_conv}. We also investigate with initial cosine transformation, followed by the integration of a convolution operator (Heracles-Cosine-conv-series), and further investigation with parallel architecture using cosine transformation and convolution operator to attain optimal results (Heracles-C). Our findings parallel architecture to capture both gobal and local information with  superior performance for vision transformer.


\subsection{Initial Attention Layer vs Spectral Layer vs Initial Convolutional}
We have tried including initial attention layers with later spectral layers(Heracles-C-S-inverse) as well as initial spectral layers with later attention layers  (Heracles-C-S) as shown in table-\ref{tab:init_conv_atten}. Another architecture includes interleaving attention and spectral layers alternately (Heracles-C-S-Alternate). We also tried with the initial convolution layer and later attention layer(Heracles-C-S-init-conv).  We found that initial spectral layers (which include cosine transformation + convolutional layers) followed by deeper attention layers perform optimally. A table is given below which captures the essence of these architectural combinations. Neural Architecture Search may be a great option to explore if the computation budget for the same is available. Currently, we have explored extensively with the computation available to us.

\section{Analysis and Trade-off}
 We visualize the original image and the frequency spectrum after applying both real (Cosine and Hartley) and complex (Fourier) transformations. We observed the differences in energy concentration and the distribution of frequency components in the transformed domains as shown in figure-\ref{fig_filter} 
 
To show the image energy distribution for both the real and complex transformations, we can visualize the cumulative energy of the transformed coefficients. This involves sorting the coefficients in descending order of their magnitudes and plotting the cumulative energy. Higher cumulative energy indicates that a smaller number of coefficients capture a larger proportion of the signal energy.  We illustrate the percentage of total energy captured by the top N coefficients for both real (Cosine and Hartley) and complex (Fourier) transformations and visualizes the energy concentration in a bar plot. The resulting plots will show how quickly the energy is accumulated as you consider more coefficients, providing insights into the energy concentration and sparsity of the transformed domains.

To visualize the energy compaction properties of real over complex transformations, we show the cumulative energy for increasing numbers of coefficients and observed that how quickly the cumulative energy increases for DCT compared to DFT, indicating the better energy compaction properties of DCT. The faster increase means that a smaller number of DCT coefficients can capture a larger proportion of the total energy, making Heracles amenable to model compression.

\begin{figure}[]%
\centering
\underline{Filter Characteristic of Tranformer}
\includegraphics[width=0.949\textwidth]{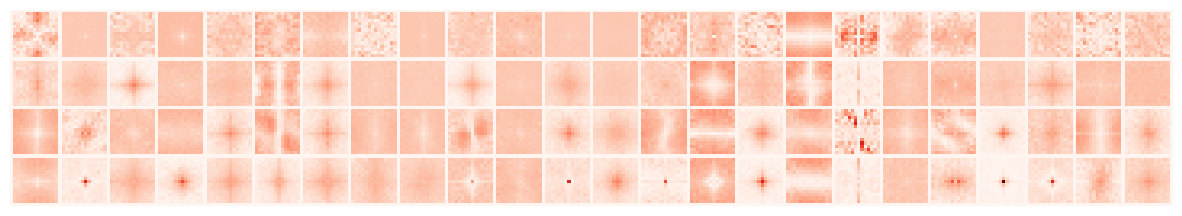}\\
\underline{Filter Characteristic of Heracles(SSM)}
\includegraphics[width=0.949\textwidth]{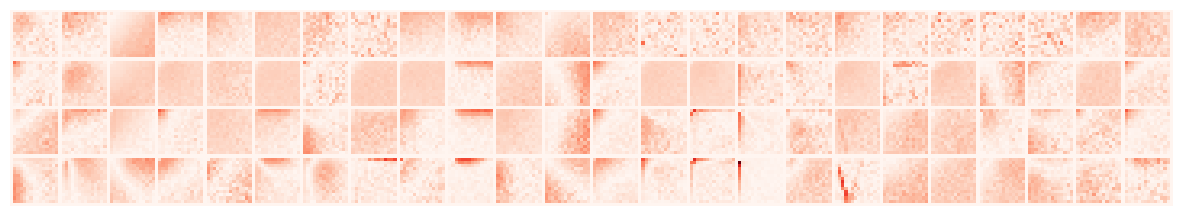}
\vspace{-0.1in}
\caption{Filter Characterisation: This figure shows the filter characterization of the initial four layers of the GFNet~\cite{rao2021global} and Heracles-C model. It clearly shows that  most of the information in Heracles-C is concentrated in low-frequency regions of an Image}\label{fig:Heracles_filter}
\vspace{-0.2in}
\end{figure}



\begin{figure}[]%
\centering
\includegraphics[width=0.949\textwidth]{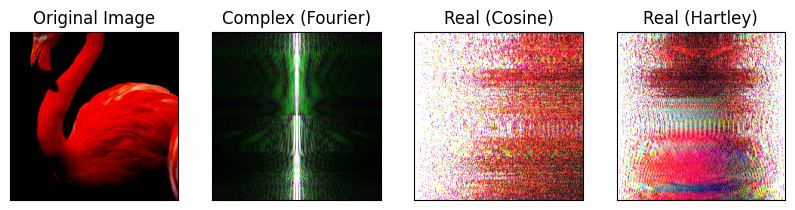}\\
\includegraphics[width=0.424\textwidth]{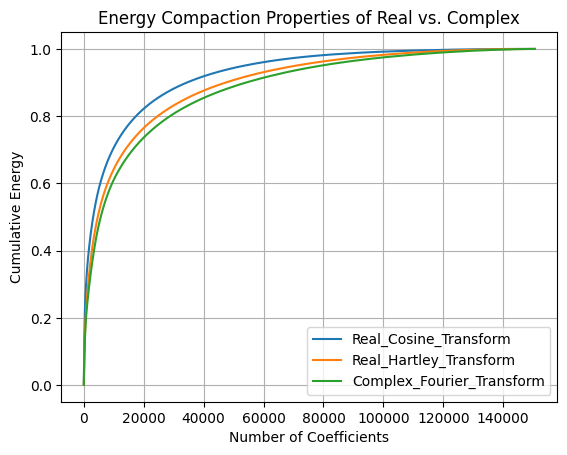}
\includegraphics[width=0.424\textwidth]{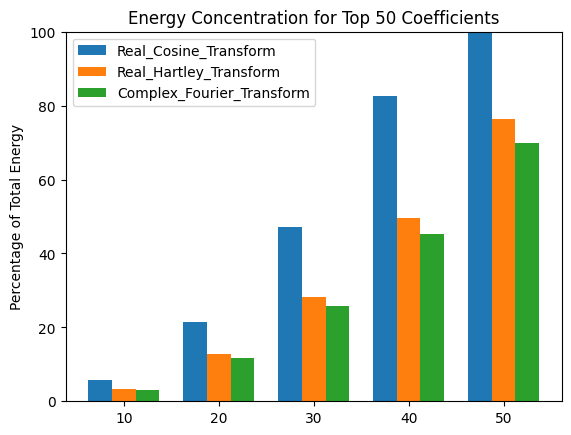}
\vspace{-0.2in}
\caption{Here we compare the spectrum of real transforms such as cosine and hartley with complex Fourier transform.  We show the energy compaction and concentration property of the real transformer over complex transforms of an Image. This shows that Heracles is amenable to model compression.}\label{fig_filter}
\vspace{-0.2in}
\end{figure}

\section{Experiments on Time Series Datasets}
\subsection{Datasets} 

\begin{table*}[htbp!]
\centering
\scalebox{0.8}{
\begin{tabular}{c|cccccc}
\toprule  
Datasets & Weather & Electricity & ETTh1 & ETTh2 & ETTm1 & ETTm2 \\
\midrule  
Features & 21 &  321 & 7 & 7 & 7 & 7 \\
Timesteps & 52696 &  26304 &  17420 & 17420 & 69680 & 69680 \\
\bottomrule 
\end{tabular}}
\caption{Statistics of popular datasets for the benchmark.}
\label{tab:data}
\end{table*}

To evaluate HTST, we $8$ popular multivariate datasets provided in \cite{autoformer} for forecasting and representation learning. \textit{Weather}\footnote{https://www.bgc-jena.mpg.de/wetter/} dataset collects 21 meteorological indicators in Germany, such as humidity and air temperature. \textit{Traffic}\footnote{https://pems.dot.ca.gov/} dataset records the road occupancy rates from different sensors on San Francisco freeways. \textit{Electricity}\footnote{https://archive.ics.uci.edu/ml/datasets/ElectricityLoadDiagrams20112014} is a dataset that describes 321 customers' hourly electricity consumption. \textit{ILI}\footnote{https://gis.cdc.gov/grasp/fluview/fluportaldashboard.html} dataset collects the number of patients and influenza-like illness ratio in a weekly frequency. \textit{ETT}\footnote{https://github.com/zhouhaoyi/ETDataset} (Electricity Transformer Temperature) datasets are collected from two different electric transformers labelled with 1 and 2, and each of them contains two different resolutions (15 minutes and 1 hour) denoted with m and h. Thus, in total we have 4 ETT datasets: \textit{ETTm1}, \textit{ETTm2}, \textit{ETTh1}, and \textit{ETTh2}. 

\subsection{Experimental settings for time series datasets}
\label{append::baseline}

We now describe the experimental setting of using Heracles for time series datasets. The default look-back windows for different baseline models could be different. For Transformer-based models, the default look-back window is $L=96$; for DLinear, the default look-back window is $L=336$. This difference is because Transformer-based baselines are easy to overfit when the look-back window is long, while DLinear tends to underfit. However, this can lead to an underestimation of the baselines. To address this issue, we re-run FEDformer, Autoformer and Informer by ourselves for six different look-back windows $L\in \{24, 48, 96, 192, 336, 720\}$. We choose the best one from those six results for each forecasting task (aka each extra prediction length on each dataset). 

Time series has been an ancient field of study, with many traditional models developed, for example, the famous ARIMA model \cite{arima}. With the bloom of the deep learning community, many new models were proposed for sequence modelling and time series forecasting before Transformer appears, such as LSTM \cite{lstm}, TCN \cite{tcn} and DeepAR \cite{deepar}. However, they are demonstrated to be not as effective as Transformer-based models in long-term forecasting tasks \cite{informer,autoformer}. Thus, we don't include them in our baselines.

We choose the SOTA Transformer-based models, including FEDformer \cite{fedformer}, Autoformer \cite{autoformer}, Informer \cite{informer}, Pyraformer \cite{pyraformer}, LogTrans \cite{logtrans}, and a recent non-Transformer-based model DLinear \cite{dlinear} as our baselines. All of the models follow the same experimental setup with prediction length and $T\in \{96, 192, 336, 720\}$. We collect baseline results from \cite{dlinear} with the default look-back window $L=96$ for Transformer-based models and $L=336$ for DLinear. But to avoid under-estimating the baselines, we also run FEDformer, Autoformer and Informer for six different look-back windows $L\in \{24, 48, 96, 192, 336, 720\}$, and always choose the best results to create strong baselines. We calculate the MSE and MAE of multivariate time series forecasting as metrics. 


\begin{figure*}[htb]
    \centering
    \includegraphics[width=0.49043\textwidth]{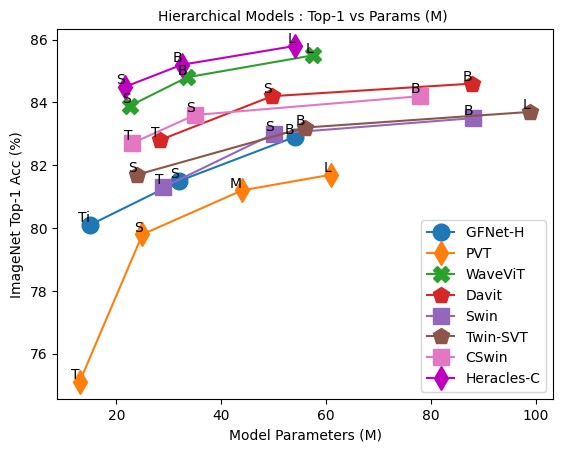}
    \includegraphics[width=0.49043\textwidth]{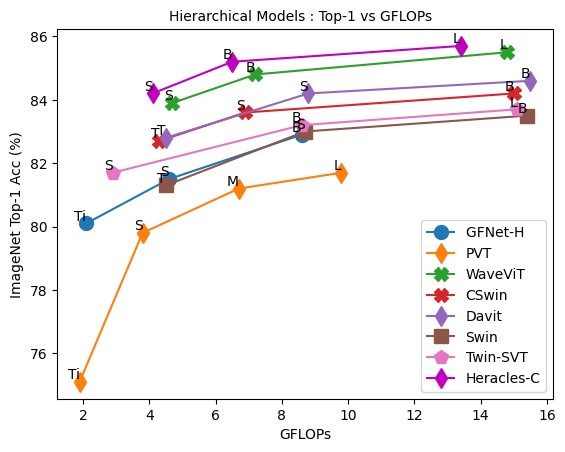}
        \caption{Comparison of ImageNet Top-1 Accuracy (\%) vs Parameters (M)  and Accuracy (\%) vs GFLOPs of various models in  Hierarchical architecture.}
        \label{fig:param}
\end{figure*}

\begin{table*}[htb]
  \centering
 \caption{\textbf{Latency(Speed test):} This table shows the Latency (mili sec) of SVT compared with Conv type network, attention type transformer, POOl type, MLP type, and Spectral type transformer. We adopt the latency table from EfficientFormer~\cite{li2022efficientformer}.}\label{tab:latency}
\begin{tabular}{lccccc}

\hline
Model & Type & Params &  {GMAC}  &  {Top-1}   & {Latency} \\
 &  & (M) &  (G)  &  (\%)   & (ms) \\
\hline

ResNet50\cite{he2016deep}& Convolution& 25.5& 4.1&  78.5& 9.0\\\hline

DeiT-S\cite{touvron2021training}& Attention& 22.5& 4.5& 81.2 &15.5\\
PVT-S\cite{wang2022pvt}& Attention& 24.5& 3.8&  79.8 & 23.8\\
T2T-14\cite{}& Attention& 21.5 &4.8&  81.5& 21.0\\
Swin-T\cite{liu2022swin}  & Attention& 29.0& 4.5&  81.3& 22.0\\
CSwin-T\cite{dong2022cswin}& Attention &23.0& 4.3&  82.7 & 28.7\\\hline
PoolFormer\cite{yu2022metaformer}& Pool& 31.0& 5.2 &81.4& 41.2\\
ResMLP-S\cite{touvron2022resmlp} &MLP &30.0& 6.0& 79.4 & 17.4\\
EfficientFormer~\cite{li2022efficientformer} & MetaBlock &31.3 &3.9& 82.4 & 13.9\\
\hline
GFNet-H-S\cite{rao2021global}& Spectral &  32.0 & 4.6 & 81.5 & 14.3\\
\rowcolor{gray!15}Heracles-C-S& Spectral & 21.7 & 4.1 & 84.5 &14.5\\
\hline
\end{tabular}
\vspace{-0.2in}
\end{table*}%

\begin{table*}[!tb]
\centering
\caption{Detailed architecture specifications for three variants of our Heracles-C with different model sizes, \emph{i.e.}, Heracles-C-S (small size), Heracles-C-B (base size), and Heracles-C-L (large size). $E_i$, $G_i$, $H_i$, and $C_i$ represent the expansion ratio of the feed-forward layer, the spectral gating number, the head number, and the channel dimension in each stage $i$, respectively.}
\begin{tabular}{c|c|c|c|c}
\Xhline{2\arrayrulewidth}
        & O\/P Size & Heracles-C-H-S & Heracles-C-H-B & Heracles-C-H-L \\ \hline
\rowcolor{gray!15}Stage 1 & $\frac{H}{4} \times \frac{W}{4}$
        & $\left[ \begin{array}{c}  E_1=8 \\ G_1=1 \\ C_1=64  \end{array} \right] \!\times\! 3$
        & $\left[ \begin{array}{c}  E_1=8 \\ G_1=1 \\ C_1=64  \end{array} \right] \!\times\! 3$
        & $\left[ \begin{array}{c}  E_1=8 \\ G_1=1 \\ C_1=96  \end{array} \right] \!\times\! 3$
        \\ \hline
\rowcolor{gray!15}Stage 2 & $\frac{H}{8} \times \frac{W}{8}$
        & $\left[ \begin{array}{c} E_2=8 \\ G_2=1 \\  C_2=128 \end{array} \right] \!\times\! 4$ 
        & $\left[ \begin{array}{c} E_2=8 \\ G_2=1 \\  C_2=128 \end{array} \right] \!\times\! 4$
        & $\left[ \begin{array}{c} E_2=8 \\ G_2=1 \\  C_2=192 \end{array} \right] \!\times\! 6$
        \\ \hline
Stage 3 & $\frac{H}{16} \times \frac{W}{16}$
        & $\left[ \begin{array}{c}  E_3=4 \\ H_3=10 \\ C_3=320 \end{array} \right] \!\times\! 6$
        & $\left[ \begin{array}{c}  E_3=4 \\ H_3=10 \\ C_3=320 \end{array} \right] \!\times\! 12$
        & $\left[ \begin{array}{c}  E_3=4 \\ H_3=12 \\ C_3=384 \end{array} \right] \!\times\! 18$
        \\ \hline
Stage 4 & $\frac{H}{32} \times \frac{W}{32}$
        & $\left[ \begin{array}{c} E_4=4 \\ H_4=14 \\ C_4=448 \end{array} \right] \!\times\! 3$
        & $\left[ \begin{array}{c} E_4=4 \\ H_4=16 \\ C_4=512 \end{array} \right] \!\times\! 3$
        & $\left[ \begin{array}{c} E_4=4 \\ H_4=16 \\ C_4=512 \end{array} \right] \!\times\! 3$
        \\ \Xhline{2\arrayrulewidth}
\end{tabular}
\label{tab:architecture}
\end{table*}

\begin{table*}[htb]
  \centering
\caption{In this table, we present a comprehensive overview of different versions of Heracles-C within the vanilla transformer architecture. The table includes detailed configurations such as the number of heads, embedding dimensions, the number of layers, and the training resolution for each variant. For Heracles-C-H models with a hierarchical structure, we refer readers to Table-\ref{tab:arch} in the main paper, which outlines the specifications for all four stages. Additionally, the table provides FLOPs (floating-point operations) calculations for input sizes of both $224\times 224$ and $384\times 384$. In the vanilla Heracles-C architecture, we utilize four spectral layers with $\alpha=4$, while the remaining attention layers are $(L-\alpha)$.} \vspace{5pt}
\setlength{\tabcolsep}{3.0pt}
    \begin{tabular}{c|c|c|c|c|c|c}
    \toprule
    Model & \#Layers &  \#heads &  \#Embedding Dim & Params (M) & Training Resolution & FLOPs (G)  \\ \midrule
    Heracles-C-Ti &  12 & 4& 256  & 9 & 224 & 1.8\\
    Heracles-C-XS &  12& 6 & 384  & 20& 224 & 4.0\\
    Heracles-C-S &  19 & 6& 384  & 32& 224 & 6.6\\
    Heracles-C-B &  19 & 8& 512  & 57 & 224& 11.5\\
      \midrule
    \rowcolor{gray!15}Heracles-C-XS &  12& 6 & 384  & 21& 384 & 13.1\\
    \rowcolor{gray!15}Heracles-C-S &  19 & 6& 384  & 33& 384 & 22.0\\
    \rowcolor{gray!15}Heracles-C-B &  19 & 8& 512  & 57& 384& 37.3\\
      \bottomrule
    \end{tabular}%
  \label{tab:arch} 
\end{table*}%

\begin{table*}[]
  \centering
  \caption{This table presents information about datasets used for transfer learning. It includes the size of the training and test sets, as well as the number of categories included in each dataset.  }
\setlength{\tabcolsep}{4.0pt}
    \begin{tabular}{c| c|c|c|c}
    \toprule
    Dataset  &   CIFAR-10   & CIFAR-100 & Flowers-102 & Stanford Cars \\\midrule
    Train Size  & 50,000  & 50,000 & 8,144 & 2,040\\
    Test Size  & 10,000  & 10,000 & 8,041  & 6,149\\
    \#Categories & 10& 100 & 196 & 102\\
     \bottomrule
    \end{tabular}%
 \vspace{-1.12em}
  \label{tab:transfer_learning_dataset}%
\end{table*}%

\begin{table*}[htb]
  \centering
\caption{We conducted a comparison of various transformer-style architectures for image classification on ImageNet. This includes \textbf{vision transformers~\cite{touvron2021training}, MLP-like models~\cite{touvron2022resmlp,liu2021pay}, spectral transformers ~\cite{rao2021global} and our Heracles-C models}, which have similar numbers of parameters and FLOPs. The top-1 accuracy on ImageNet's validation set, as well as the number of parameters and FLOPs, are reported. All models were trained using $224 \times 224$ images. We used the notation "↑384" to indicate models fine-tuned on $384 \times 384$ images for 30 epochs. }\vspace{5pt}
\setlength{\tabcolsep}{3.0pt}
    \begin{tabular}{lcccccc}\toprule
    Model & Params (M) & FLOPs (G) & Resolution & Top-1 Acc. (\%) & Top-5  Acc.  (\%) \\ \midrule
    gMLP-Ti~\cite{liu2021pay} & 6     & 1.4   & 224   & 72.0  & - \\
     DeiT-Ti~\cite{touvron2021training} & 5    & 1.2  & 224   & 72.2 &  91.1  \\     
      GFNet-Ti~\cite{rao2021global}  & 7 & 1.3 & 224  & 74.6 & 92.2 \\
     \rowcolor{gray!15} Heracles-C-T & 9 & 1.8 & 224 &  76.9 & 93.4\\ 
     \midrule
     ResMLP-12~\cite{touvron2022resmlp} & 15    & 3.0   & 224   & 76.6  & - \\
      GFNet-XS~\cite{rao2021global} & 16 & 2.9 & 224 & 78.6 & 94.2\\ 
     
\rowcolor{gray!15} Heracles-C-XS & 20& 4.0& 224& 79.9 &94.5\\\midrule
     DeiT-S~\cite{touvron2021training} & 22    & 4.6   & 224   & 79.8  & 95.0  \\
     gMLP-S~\cite{liu2021pay} & 20    & 4.5   & 224   & 79.4  & - \\
     GFNet-S~\cite{rao2021global} & 25 & 4.5 & 224 & 80.0 & 94.9\\ 
     
 \rowcolor{gray!15}Heracles-C-S & 32& 6.6& 224& 81.5 &95.3\\\midrule
     ResMLP-36~\cite{touvron2022resmlp} & 45    & 8.9   & 224   & 79.7  & - \\
     GFNet-B~\cite{rao2021global} & 43 & 7.9 & 224 & 80.7 & 95.1 \\ 
     gMLP-B~\cite{liu2021pay} & 73    & 15.8  & 224   & 81.6  & - \\
    DeiT-B~\cite{touvron2021training} & 86    & 17.5  & 224   & 81.8  & 95.6 \\
     
    \rowcolor{gray!15}  Heracles-C-B &57& 11.6& 224 &\textbf{82.0}& \textbf{95.6}\\ 
     \midrule

    GFNet-XS↑384~\cite{rao2021global} & 18 & 8.4& 384 &  80.6 & 95.4\\ 
    GFNet-S↑384 ~\cite{rao2021global}& 28 &  13.2 & 384 & 81.7 & 95.8\\
    GFNet-B↑384 ~\cite{rao2021global} & 47 & 23.3 & 384 & 82.1 & 95.8\\ 
    \rowcolor{gray!15} Heracles-C-XS↑384 & 21 & 13.1& 384 &  82.2 & 95.8\\ 
     \rowcolor{gray!15} Heracles-C-S↑384 & 33 &  22.0 & 384 & 83.1 & 96.4\\
     \rowcolor{gray!15} Heracles-C-B↑384  & 57 & 37.3 & 384 & 83.0 & 96.2\\ \bottomrule
    \end{tabular}%
  \label{tab:finetune} 
\end{table*}%

\section{Dataset and Training Details:}


\subsection{Latency Analysis}


It's important to highlight that Fourier Transforms, as mentioned in the GFNet\cite{rao2021global}, are not inherently capable of performing low-pass and high-pass separations. In contrast, GFNet consistently employs tensor multiplication, a method that, while effective, may be less efficient compared to Einstein multiplication. The latter approach is known for reducing the number of parameters and computational complexity. As a result, Heracles-C does not lag behind in terms of performance or computational complexity; rather, it gains enhanced representational power. This is exemplified in Table~\ref{tab:latency}, which provides a comparison of latency, FLOPS (Floating-Point Operations per Second), and the number of parameters. Table~\ref{tab:latency} specifically demonstrates the latency (measured in milliseconds) of Heracles-C in relation to various network types, including convolution-based networks, Attention-based Transformer networks, Pool-based Transformer networks, MLP-based Transformer networks, and Spectral-based Transformer networks. The reported latency values are on a per-sample basis, measured on an A100 GPU.


\subsection{Dataset and Training Setups  on ImageNet-1K for Image Classification task}

In this section, we outline the dataset and training setups for the Image Classification task on the ImageNet-1K benchmark dataset. The dataset comprises 1.28 million training images and 50K validation images, spanning across 1,000 categories. To train the vision backbones from scratch, we employ several data augmentation techniques, including RandAug, CutOut, and Token Labeling objectives with MixToken. These augmentation techniques help enhance the model's generalization capabilities. For performance evaluation, we measure the trained backbones' top-1 and top-5 accuracies on the validation set, providing a comprehensive assessment of the model's classification capabilities. In the optimization process, we adopt the AdamW optimizer with a momentum of 0.9, combining it with a 10-epoch linear warm-up phase and a subsequent 310-epoch cosine decay learning rate scheduler. These strategies aid in achieving stable and effective model training. To handle the computational load, we distribute the training process on 8 V100 GPUs, utilizing a batch size of 128. This distributed setup helps accelerate the training process while making efficient use of available hardware resources.  The learning rate and weight decay are fixed at 0.00001 and 0.05, respectively, maintaining stable training and mitigating overfitting risks. 

\subsection{Training setup for Transfer Learning}

In the context of transfer learning, we sought to evaluate the efficacy of our vanilla Heracles-C architecture on widely-used benchmark datasets, namely CIFAR-10~\cite{krizhevsky2009learning}, CIFAR100~\cite{krizhevsky2009learning}, Oxford-IIIT-Flower~\cite{nilsback2008automated} and Standford Cars~\cite{krause20133d}.  Our approach followed the methodology of previous studies~\cite{tan2019efficientnet,dosovitskiy2020image,touvron2021training,touvron2022resmlp,rao2021global}, where we initialized the model with pre-trained weights from ImageNet and subsequently fine-tuned it on the new datasets.

Table-4 in the main paper presents a comprehensive comparison of the transfer learning performance of both our basic and best models against state-of-the-art CNNs and vision transformers. To maintain consistency, we employed a batch size of 64, a learning rate (lr) of 0.0001, a weight-decay of 1e-4, a clip-grad value of 1, and performed 5 epochs of warmup. For the transfer learning process, we utilized a pre-trained model that was initially trained on the ImageNet-1K dataset. This pre-trained model was fine-tuned on the specific transfer learning dataset mentioned in Table-\ref{tab:transfer_learning_dataset} for a total of 1000 epochs.


\subsection{Training setup for Task Learning}


In this section, we conduct an in-depth analysis of the pre-trained Heracles-C-H-small model's performance on the COCO dataset for two distinct downstream tasks involving object localization, ranging from bounding-box level to pixel level. Specifically, we evaluate our Heracles-C-H-small model on instance segmentation tasks, such as Mask R-CNN~\cite{he2017mask}, as demonstrated in Table-5 of the main paper.

For the downstream task, we replace the CNN backbones in the respective detectors with our pre-trained Heracles-C-H-small model to evaluate its effectiveness. Prior to this, we pre-train each vision backbone on the ImageNet-1K dataset, initializing the newly added layers with Xavier initialization \cite{glorot2010understanding}. Next, we adhere to the standard setups defined in \cite{liu2021swin} to train all models on the COCO train2017 dataset, which comprises approximately 118,000 images. The training process is performed with a batch size of 16, and we utilize the AdamW optimizer \cite{loshchilovdecoupled} with a weight decay of 0.05, an initial learning rate of 0.0001, and betas set to (0.9, 0.999). To manage the learning rate during training, we adopt the step learning rate policy with linear warm-up at every 500 iterations and a warm-up ratio of 0.001. These learning rate configurations aid in optimizing the model's performance and convergence.



\subsection{Training setup for Fine-tuning task}

In our main experiments, we conduct image classification tasks on the widely-used ImageNet dataset \cite{deng2009imagenet}, a standard benchmark for large-scale image classification. To ensure a fair and meaningful comparison with previous research \cite{touvron2021training, touvron2022resmlp, rao2021global}, we adopt the same training details for our Heracles-C models. For the vanilla transformer architecture (Heracles-C), we utilize the hyperparameters recommended by the GFNet implementation \cite{rao2021global}. Similarly, for the hierarchical architecture (Heracles-C-H), we employ the hyperparameters recommended by the WaveVit implementation \cite{yao2022wave}. During fine-tuning at higher resolutions, we follow the hyperparameters suggested by the GFNet implementation \cite{rao2021global} and train our models for 30 epochs.

All model training is performed on a single machine equipped with 8 V100 GPUs. In our experiments, we specifically compare the fine-tuning performance of our models with GFNet \cite{rao2021global}. Our observations indicate that our Heracles-C models outperform GFNet's base spectral network. For instance, Heracles-C-S(384) achieves an impressive accuracy of 83.0\%, surpassing GFNet-S(384) by 1.2\%, as presented in Table~\ref{tab:finetune}. Similarly, Heracles-C-XS and Heracles-C-B outperform GFNet-XS and GFNet-B, respectively, highlighting the superior performance of our Heracles-C models in the fine-tuning process.

\end{document}